\documentclass[5p]{elsarticle}
\usepackage{lineno}
\usepackage{enumerate}
\usepackage{graphicx}
\usepackage{amsmath,amsfonts,amssymb}
\usepackage{gensymb}
\usepackage{bm}
\usepackage{diagbox}
\usepackage{tabularx}
\usepackage{rotating}
\newcolumntype{L}[1]{>{\raggedright\let\newline\\\arraybackslash\hspace{0pt}}m{#1}}
\newcolumntype{C}[1]{>{\centering\arraybackslash}m{#1}}
\newcolumntype{R}[1]{>{\raggedleft\let\newline\\\arraybackslash\hspace{0pt}}m{#1}}

\modulolinenumbers[5]

\journal{Computer Methods and Programs in Biomedicine}

\begin{document}

\begin{frontmatter}

\title{Single-view 2D CNNs with fully automatic non-nodule categorization for false positive reduction in pulmonary nodule detection}

\author[address]{Hyunjun Eun}

\author[address]{Daeyeong Kim}
\author[address2]{Chanho Jung}

\author[address]{Changick Kim\corref{mycorrespondingauthor}}
\cortext[correspondingauthor]{Corresponding author}
\ead{changick@kaist.ac.kr}

\address[address]{School of Electrical Engineering, Korea Advanced Institute of Science and Technology, Republic of Korea}
\address[address2]{Department of Electrical Engineering, Hanbat National University, Republic of Korea}

\begin{abstract}
\textbf{Background and Objective:}
In pulmonary nodule detection, the first stage, candidate detection, aims to detect suspicious pulmonary nodules.
However, detected candidates include many false positives and thus in the following stage, false positive reduction, such false positives are reliably reduced.
Note that this task is challenging due to 1) the imbalance between the numbers of nodules and non-nodules and 2) the intra-class diversity of non-nodules.
Although techniques using 3D convolutional neural networks (CNNs) have shown promising performance, they suffer from high computational complexity which hinders constructing deep networks.
To efficiently address these problems, we propose a novel framework using the ensemble of 2D CNNs using single views, which outperforms existing 3D CNN-based methods.

\noindent \textbf{Methods:}
Our ensemble of 2D CNNs utilizes single-view 2D patches to improve both computational and memory efficiency compared to previous techniques exploiting 3D CNNs.
We first categorize non-nodules on the basis of features encoded by an autoencoder.
Then, all 2D CNNs are trained by using the same nodule samples, but with different types of non-nodules.
By extending the learning capability, this training scheme resolves difficulties of extracting representative features from non-nodules with large appearance variations.
Note that, instead of manual categorization requiring the heavy workload of radiologists, we propose to automatically categorize non-nodules based on the autoencoder and k-means clustering.

\noindent \textbf{Results:}
We performed extensive experiments to validate the effectiveness of our framework based on the database of the lung nodule analysis 2016 challenge.
The superiority of our framework is demonstrated through comparing the performance of five frameworks trained with differently constructed training sets.
Our proposed framework achieved state-of-the-art performance (0.922 of the competition performance metric score) with low computational demands (789K of parameters and 1,024M of floating point operations per second).

\noindent \textbf{Conclusion:}
We presented a novel false positive reduction framework, the ensemble of single-view 2D CNNs with fully automatic non-nodule categorization, for pulmonary nodule detection.
Unlike previous 3D CNN-based frameworks, we utilized 2D CNNs using 2D single views to improve computational efficiency.
Also, our training scheme using categorized non-nodules, extends the learning capability of representative features of different non-nodules.
Our framework achieved state-of-the-art performance with low computational complexity.
\end{abstract}


\end{frontmatter}


\section{Introduction}
One of the most common causes of cancer death is lung cancer all over the world \cite{valentinaEJR2012}.
Lung cancer accounts for 27\% of all cancer deaths with estimated 224,390 new cases and 158,080 deaths in the United States alone in 2016 \cite{rebeccalCJC2016}.
Low-dose computed tomography (CT) is widely used in screening programs to reduce this high mortality rate.
The detection and the treatment of small nodules at an early stage are important to prevent lung cancer deaths.
The low-dose CT enables the detection of small nodules and carcinoma at an early stage \cite{kanekoRadio1996, miettinenRadio2011}.
Moreover, screening with the low-dose CT can reduce the mortality of lung cancer by 20\%, as compared to chest radiography \cite{tnlctrt2011NEJM}.
For this reason, many low dose CT imaging-based screening programs \cite{swensenRadio2003,henschkeRadio2004} have been developed for efficient detection of lung cancer, but the reading process in the screening programs leads to a heavy workload for radiologists.
Furthermore, this heavy workload may cause errors in interpretation and detection of nodules \cite{gurneyRadio1996,liRadio2002}.

Computer-aided detection (CAD) systems can alleviate the above problems in human screening by assisting radiologists.
Specifically, CAD systems improve accuracy of cancer detection and reduce the work for radiologists in exam evaluation \cite{firminoBEO2014}.
The CAD systems for detecting pulmonary nodules also have been investigated and refined for over a decade, but these are still rarely employed in clinical practice because of two main reasons \cite{firminoBEO2014}.
First, nodules and non-nodules are misclassified since nodules have a wide variation in size, shape, and contextual environments and non-nodules also have diverse patterns \cite{douTBME2016}.
Since the CAD systems generally pursue a very high sensitivity (i.e., true positive rate) as the priority not to miss pulmonary nodules, the number of misclassified nodules can be reduced.
However, a large number of false positives (i.e., a misclassified non-nodule as a nodule) occur at the cost of high sensitivity.
These numerous false positives that require extra attention from radiologists also hinder the practical use of CAD systems \cite{rubinJTI2015}.
Second, the use of volumetric 3D images requires very high computational cost and huge memory storage, which directly affect the speed of the training and detection processes.

Most CAD systems for pulmonary nodule detection adopt a two-stage framework including candidate detection and false positive reduction.
In the first stage, nodule candidates are detected to include as many malignant nodules as possible based on simple approaches \cite{murphyMIA2009,jacobsMIA2014,setioMP2014}.
The goal of this stage is to detect all suspicious lesions although it returns a large number of false positives.
Murphy et al. \cite{murphyMIA2009} compute the shape index and the curvedness as 3D nodule features.
Thresholding on these two features is applied to define seed points. Then, candidate nodules are defined via clustering and merging of the seed points based on their locations.
Jacobs et al. \cite{jacobsMIA2014} apply the double-threshold density mask to obtain rough nodule voxels.
Connected component analysis is performed to cluster the rough nodule voxels as nodules after morphological erosion for separating nodules with edges of the lung, vessels, and airways.
Setio et al. \cite{setioMP2014} propose the candidate detection that consists of initial candidate detection, clustering, and refinement.
A multistage process of thresholding and morphological opening is performed to detect initial candidates.
Then, all connected voxels are clustered to form candidates by connected component analysis.
Region growing and morphological operation are further used to obtain more accurate coordinates of the candidates.

Since candidate detection provides a large number of nodule candidates, false positives need to be reduced in the following stage.
To this end, several machine learning based techniques \cite{jacobsMIA2014,yeTBME2009,armatoR1999,valenteCMPB2016} have been proposed.
These methods utilize diverse hand-crafted features from low-level to high-level.
Intensity, color, and volume size would be used as low-level features.
For high-level features, compactness, texture, shape, elongation, and curvature tensor are introduced.
However, these hand-crafted features need to be carefully designed for successfully training classifiers.
Neural networks such as artificial neural networks (ANNs) and convolutional neural networks (CNNs) can eliminate the work of the designing features.
With the recent successes of convolutional neural networks (CNNs) in computer vision, several medical imaging applications \cite{shiMTA2018,ciompiMIA2015,setioTMI2016,douTBME2016,suzukiRPT2017,zhangBSPC2018} also widely employ them for false positive reduction in pulmonary nodule detection.

\begin{figure}[t!]
\footnotesize
\begin{minipage}[t]{0.99\linewidth}
\centering{\includegraphics[width=.9\linewidth]{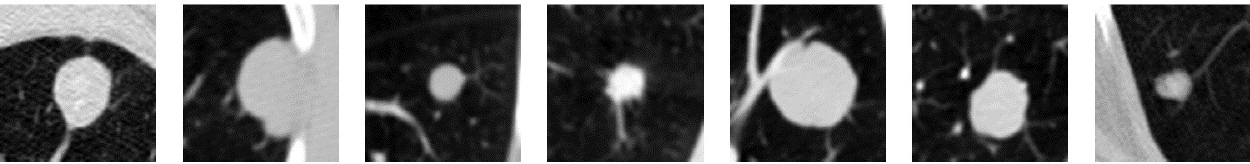}} \\
\centering{(a)} \\
\centering{\includegraphics[width=.9\linewidth]{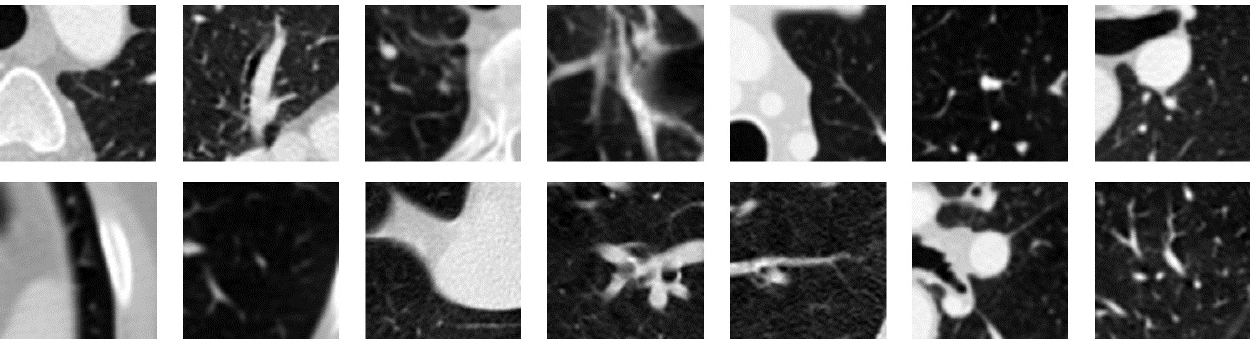}} \\
\centering{(b)}
\end{minipage}
\caption{\label{fig1}
Example images of nodule candidates. (a) Nodules. (b) Non-nodules with large appearance variation.}
\end{figure}

In this paper, we propose a novel framework, an ensemble of 2D CNNs using single views, for efficient and accurate false positive reduction in pulmonary nodule detection.
In contrast to previous methods exploiting 3D volumetric patches, the proposed framework utilizes 2D patches to improve memory usage and computational efficiency without a decrease in performance.
Although 3D CNNs utilize more information, it is difficult to learn representative features from 3D data without a complicated architecture consisting of a large number of parameters and layers.
However, limited computation resources lead to difficulty in constructing the complicated 3D architecture.
From successful works \cite{suzukiMP2003,tajbakhshPR2017} based on a 2D single view, we claim that representative features of nodules can be learned from the single view.
For example, round shapes of nodules are generally distinguished in the single view of a 3D CT scan as can be seen in Fig. 1 (a).
To improve a capacity of a 2D CNN for learning representative features from non-nodules with large appearance variations (see Fig. 1 (b)), we adopt the training scheme \cite{suzukiMP2003} of an ensemble network where each network is trained on different types of non-nodules while using the same nodules.
From this training scheme, we can extend the capability of each 2D CNN to distinguish the specific type of non-nodules and nodules successfully.
Furthermore, the ensemble of these 2D CNNs can decrease a large number of false positives by alleviating the bias of single network.
Unlike Suzuki et al. \cite{suzukiMP2003} which manually categorizes non-nodules to perform this training scheme, we introduce a feature extraction method using an autoencoder \cite{vincentJMLR2010} and k-means clustering \cite{macqueenBSMSP1967} for fully automatic non-nodule categorization without the hand of radiologists.
Note that non-nodules essentially indicate false positives in nodule candidates.
By using an autoencoder, we can automatically extract representative features of non-nodules.
Then we categorize non-nodules regarding these features by using k-means clustering.
This scheme effectively eliminates the work of radiologists to categorize types of non-nodules.
We performed extensive experiments on a large scale benchmark dataset.
We compared the proposed framework based on five different training schemes to confirm the effectiveness of our key ideas.
Also, we achieved the best performance compared to state-of-the-art techniques for false positive reduction in pulmonary detection.

Our contributions can be summarized as follows:

1)	We introduce a novel framework, single-view 2D CNNs with fully automatic non-nodule categorization, which achieves the higher computation efficiency and better performance of false positive reduction in pulmonary nodule detection than state-of-the-art techniques.

2)	Unlike manual non-nodule categorization \cite{suzukiMP2003}, we introduce a fully automatic non-nodule categorization by utilizing an autoencoder and k-means clustering. We can extend the learning capability of 2D CNNs by training each 2D CNN with a different category of non-nodules while using the same nodules.

The remainder of this paper is organized as follows. In Section 2, we review on previous false positive reduction approaches. In Section 3, the benchmark dataset that we used is described. We explain the details of our proposed framework in Section 4. Extensive experiments on the benchmark dataset are drawn in Section 5. Finally, we conclude this paper in Section 6.

\section{Related Work}
We first review methods using hand-crafted features to train classifiers.
From both 2D and 3D images, Ye et al. \cite{yeTBME2009} compute 15 features based on intensity, compactness, shape index, elongation, volume size, and sphericity.
In order to classify the feature vectors, a weighted SVM \cite{osuna1997} is used.
Jacobs et al. \cite{jacobsMIA2014} introduce a rich set of 128 features based on intensity, texture, shape, and context to describe sub-solid nodule candidates.
Six classifiers such as SVM-RBF \cite{vaponik1998}, k-nearest neighbors \cite{cover1TIT1967}, GentleBoost \cite{friedmanAS2000}, random forest \cite{breimanML2001}, nearest mean \cite{fukunaga1990}, and a linear discriminant classifier are utilized to compare performances of false positive reduction.
Armato et al. \cite{armatoR1999} first apply multiple gray-level thresholding to the volumetric lung regions to extract nodule candidate.
2D and 3D geometric and gray-level features for each candidate are computed.
Then, linear discriminant analysis is utilized to merge features and reduce false positives.
It is noted that, to successfully train classifiers of the aforementioned methods, it is important to carefully design hand-crafted features with much effort.

Recently, many methods utilize neural networks to eliminate the work of the designing hand-crafted features.
Suzuki et al. \cite{suzukiMP2003} introduce the multiple massive training artificial neural network (Multi-MTANN).
A single MTANN adopts raw pixel data as input to predict a nodule score of each pixel.
A Multi-MTANN consists of plural MTANNs in parallel to distinguish between nodules and diverse types of non-nodules.
In the Multi-MTANN, each MTANN deals with binary classification between nodules and a single type of non-nodules.
Suzuki et al. have constructed the Multi-MTANN with nine MTANNs, which implies 10-class classification (i.e., one type of nodules and nine types of non-nodules).
This training scheme provides better performance by increasing the capability of MTANNs.
However, the categorization of non-nodules performed by humans would require extensive workload.
According to CNNs show tremendous successes in computer vision, many approaches also widely employ CNNs for false positive reduction in pulmonary nodule detection.
Shi et al. \cite{shiMTA2018} introduce transfer learning for false positive reduction.
They tune a pre-trained CNN in ImageNet \cite{dengMTA2018} to perform a pulmonary nodule classification task.
Ciompi et al. \cite{ciompiMIA2015} perform the detection of pulmonary peri-fissural nodules using an ensemble of 2D CNNs.
The input of each 2D CNN is defined as axial, coronal, and sagittal views of a 3D candidate patch.
Also, the OverFeat method \cite{sermanetarXiv2014} is modified to construct the 2D CNN.
Setio et al. \cite{setioTMI2016} introduce multi-view convolutional networks (MVCNs) that exploit nine 2D patches as input.
The nine 2D patches are extracted from nine symmetrical planes of a 3D nodule candidate patch.
These nine 2D patches pass through each corresponding 2D CNN to generate feature vectors.
Then, the nodule likelihood of the candidate patch is derived by a mixed-fusion scheme on nine 2D CNNs.
Dou et al. \cite{douTBME2016} exploit three 3D CNNs to encode richer spatial information and extract more discriminative representations than 2D CNNs.
Three 3D CNNs are independently trained with three different sizes of 3D patches to handle the large variation of nodule sizes.
The final nodule probability is defined as a weighted sum of the outputs of three 3D CNNs.
Many teams that participated in the false positive reduction track in the lung nodule analysis 2016 (LUNA16) challenge also exploit 3D CNNs with 3D volumetric data to obtain more information compared to 2D data.
These teams rank higher than teams using 2D CNNs with 2D data.
However, employing 3D CNNs suffers from large memory and computation work.

\begin{figure*}[t!]
\footnotesize
\begin{minipage}[t]{0.6\linewidth}
\centering{\includegraphics[width=1.\linewidth]{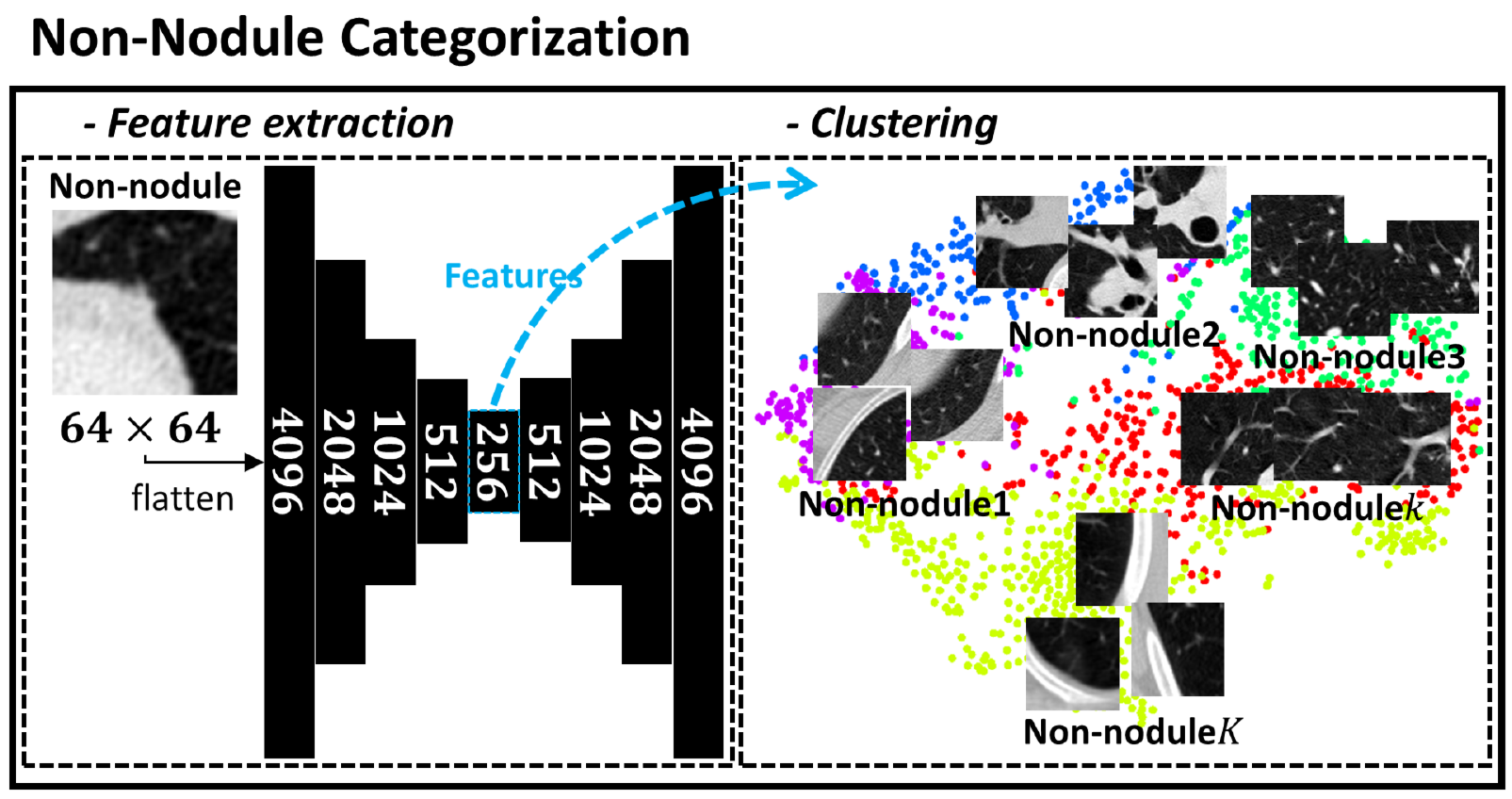}} \\
\centering{(a)} \\
\end{minipage}
\begin{minipage}[t]{0.35\linewidth}
\centering{\includegraphics[width=1.\linewidth]{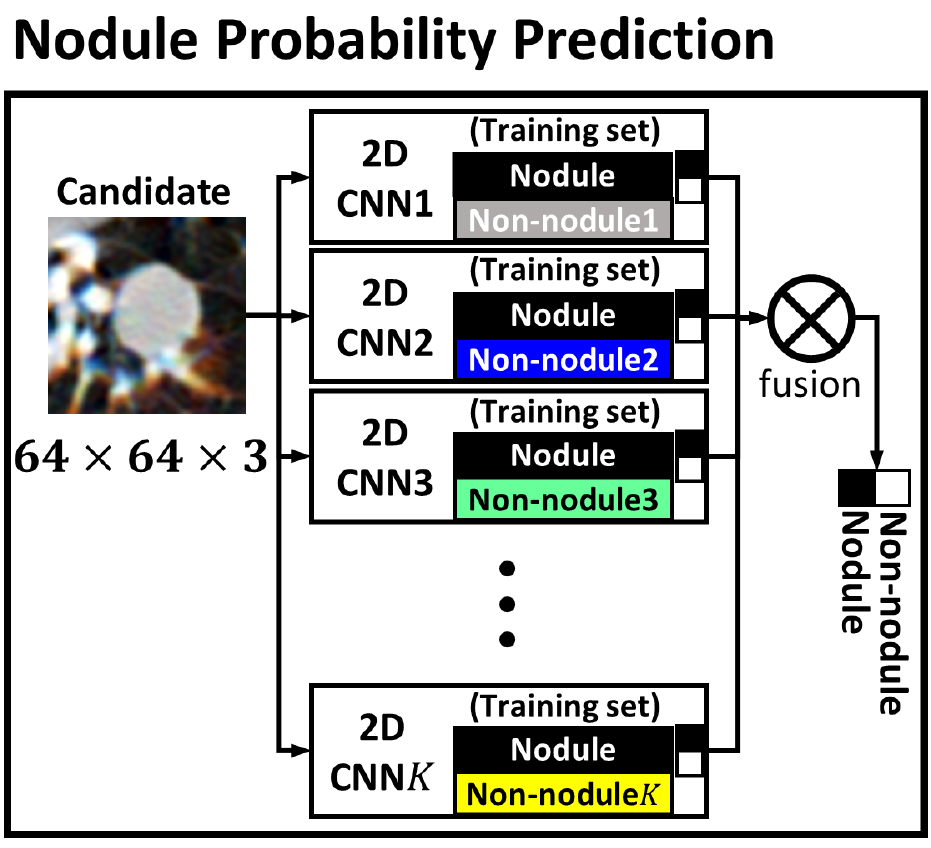}} \\
\centering{(b)} \\
\end{minipage}
\caption{\label{fig2}
Overview of the proposed framework for false positive reduction in pulmonary nodule detection. (a) Non-nodule categorization. (b) Nodule probability prediction.}
\end{figure*}
We additionally review organ segmentation methods since organ segmentation can assist false positive reduction by explicitly defining regions that possibly contain nodules. 
Organ segmentation can be integrated into false positive reduction as pre- or post-processing.
Wang et al. \cite{wangIJCARS2016} propose a fully automatic 3D segmentation framework to segment liver from abdominal CT images.
First, the similarities between all atlases and the test dataset are calculated to generate a subject-specific probabilistic atlas and determine the most likely liver region.
Then, a maximum a posteriori classification and a shape-intensity prior level set are introduces to segment liver.
Shi et al. \cite{shiMIA2017} introduce low-rank and sparse decomposition (LRSD) based shape model for automatic organ segmentation.
The shape model is initialized based on patient-speciﬁc LRSD based probabilistic atlas.
LRSD based shape model can handle non-Gaussian gross errors caused by weak and misleading appearance cues of large lesions, complex shape variations, and poor adaptation.
Linguraru et al. \cite{linguraruTMI2012} propose a parameterization of 3D surfaces for the comparison of objects by point-to-point correspondence.
After initially segmenting liver, a shape-driven geodesic active contour improves the segmentation result.

\section{Material}
We use the publicly available database provided by the LUNA16 challenge.
The database contains 888 CT scans with a slice thickness less than 2.5 $mm$.
The CT scans have 512$\times$512 resolution in the transverse plane and 0.74$\times$0.74 $mm^2$ element spacing.
These CT scans are originally from the Lung Image Database Consortium (LIDC) database \cite{armatoMP2011}.
Annotations for nodules are labeled by a two-phase manual process by four expert radiologists.
Each radiologist annotated suspicious lesions as non-nodule, nodule $<$3 $mm$, or nodule $\geq$3 $mm$.
For the LUNA16 database, all nodules $\geq$3 $mm$ accepted by more than three radiologists are selected as a reference standard.
Annotations as non-nodules, nodules $<$3 $mm$, and nodules $\geq$3 $mm$ annotated by less than two radiologists are referred to as irrelevant findings.
Additionally, nodule candidates with center locations detected by three candidate detection algorithms \cite{murphyMIA2009,jacobsMIA2014,setioMP2014} are provided for the false positive reduction track in the LUNA16 challenge.
These three algorithms detected 1,120 out of 1,186 nodules with 551,065 candidates. Note that our quantitative evaluation in Section 5 was performed based on the initial set of 1,186 nodules.
To sum up, the database of the false positive reduction track consists of 551,065 candidates with 1,351 nodules that contain multiple candidates per nodule.

\section{Method}
The proposed framework, an ensemble of single-view 2D CNNs with fully automatic non-nodule categorization, aims to achieve more efficient and accurate false positive reduction in pulmonary nodule detection.
In Fig. 2, we delineate the proposed framework composed of two parts: non-nodule categorization for training and nodule probability prediction.
We automatically categorize non-nodules by using an autoencoder and k-means clustering at the non-nodule categorization stage.
Based on these categorized non-nodules, we train each 2D CNN with a different category of non-nodules and same nodules to predict a nodule probability.
In testing, a nodule probability of a nodule candidate is predicted by combining the prediction results of each 2D CNN.

\subsection{Non-nodule categorization}
To train 2D CNNs with different types of non-nodules and same nodules, we categorize non-nodules with large appearance variations.
This training scheme can generate an expert network to distinguish a specific type of non-nodules and nodules by extending the learning capability \cite{suzukiMP2003}.
Furthermore, the ensemble of these 2D CNNs has the great ability of effectively reducing false positives as will be seen in Section 5.
For fully automatic non-nodule categorization without the hand of radiologists, we introduce an autoencoder to extract representative features of non-nodules and adopt the k-means clustering to group non-nodules based on these features.

\subsubsection{Feature extraction}
An autoencoder is one of the deep neural networks that aims to learn representative features from data.
The autoencoder consists of an input layer, an output layer, and several hidden layers.
From the input layer to the middle hidden layer, an input is encoded to representative features.
Then, the encoded features are reconstructed to have the output same as the input from the middle hidden layer to the output layer.
All adjacent layers in the autoencoder are fully-connected, which indicates that each neuron is connected with every neuron in adjacent layers.
We construct the autoencoder with seven hidden layers (see the left of Fig. 2(a)) as
\begin{eqnarray}
\bm{h}^l = \bm{\sigma}(\bm{b}^l + \bm{W}^l\bm{h}^{l-1}),
\end{eqnarray}
where $\bm{h}^{l-1}$ and $\bm{h}^{l}$ denote the input and the output feature vector of the $l$th layer, respectively.
Indeed, the output of the previous layer becomes the input of the current layer.
$\bm{W}^l$ is the weight matrix.
$\bm{b}^l$ is the bias vector.
$\bm{\sigma}(\cdot)$ is the non-linear activation function defined as the rectified linear units (ReLU) \cite{glorotICAIS2011}.

Let $\{(\bm{X}^{(1)}_{\tiny{\text{AE}}},\hat{\bm{X}}^{(1)}_{\tiny{\text{AE}}}), ..., (\bm{X}^{(M)}_{\tiny{\text{AE}}},\hat{\bm{X}}^{(M)}_{\tiny{\text{AE}}})\}$ denote a set of $M$ paired non-nodule samples, where $\bm{X}^{(m)}_{\tiny{\text{AE}}}$ is an input patch of size $64 \times 64$ extracted on an axial plane of a 3D CT scan and $\hat{\bm{X}}^{(m)}_{\tiny{\text{AE}}}$ is the corresponding reconstructed patch.  
The size of the input patch is determined by considering all nodule sizes and surrounding information.
As shown in Fig. 2(a), we flatten $\bm{X}^{(m)}_{\tiny{\text{AE}}}$ as a 1D vector $\bm{x}^{(m)}_{\tiny{\text{AE}}}$ to feed the first hidden layer (i.e., $\bm{h}^0 = \bm{x}^{(m)}_{\tiny{\text{AE}}}$).
For training the parameters $\theta_{\tiny{\text{AE}}}$ of the autoencoder, we minimize the loss function defined as
\begin{eqnarray}
\begin{aligned}
&\mathcal{L}(\theta_{\tiny{\text{AE}}}) = \frac{1}{M} \sum^M_{m=1} \lVert \bm{x}^{(m)}_{\tiny{\text{AE}}}-\bm{\hat{x}}^{(m)}_{\tiny{\text{AE}}} \rVert^2_2, \\
\end{aligned}
\end{eqnarray}
where $\bm{\hat{x}}^{(m)}_{\tiny{\text{AE}}}$ is the reconstructed vector, the output of the last layer (i.e., $\bm{\hat{x}}^{(m)}_{\tiny{\text{AE}}} = \bm{h}^8$).
We randomly initialized the weights based on Gaussian distribution $\mathcal{N}(0, 0.05^2)$.
The biases were initialized as zeros.
All parameters were updated by using the Adam optimization technique \cite{kingmaarXiv2017}.
We used the learning rate 0.001 with $4\%$ decay on every 1000 iterations.
We experimentally determined all hyper-parameters for training.
We show several reconstruction results of non-nodules in Fig. 3.
As can be seen, the trained autoencoder faithfully reconstructs non-nodules similar to input.
By using the trained autoencoder, we extract the output of the fourth layer $\bm{h}^4 \in \mathbb{R}^{256\times1}$ as a feature vector for an input non-nodule.

\begin{figure}[t!]
\footnotesize
\centering{(a)}
\centering{\includegraphics[width=0.9\linewidth]{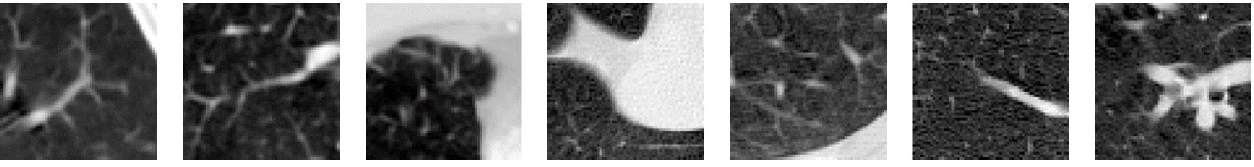}} \\
\centering{(b)}
\centering{\includegraphics[width=0.9\linewidth]{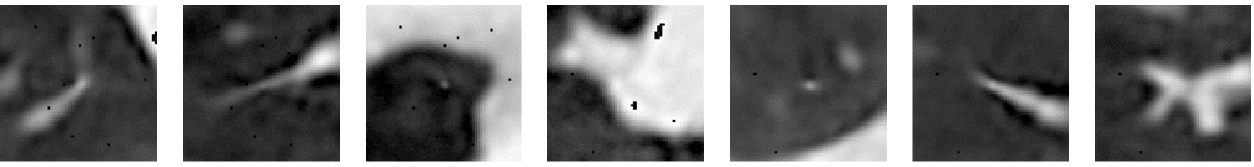}} \\
\caption{\label{fig3}
Reconstruction results of an autoencoder. (a) Input non-nodule $\bm{X}_{\tiny{\text{AE}}}$. (b) Corresponding reconstructed non-nodule $\hat{\bm{X}}_{\tiny{\text{AE}}}$.}
\end{figure}

\begin{figure}[t!]
\centering{\includegraphics[width=0.9\linewidth]{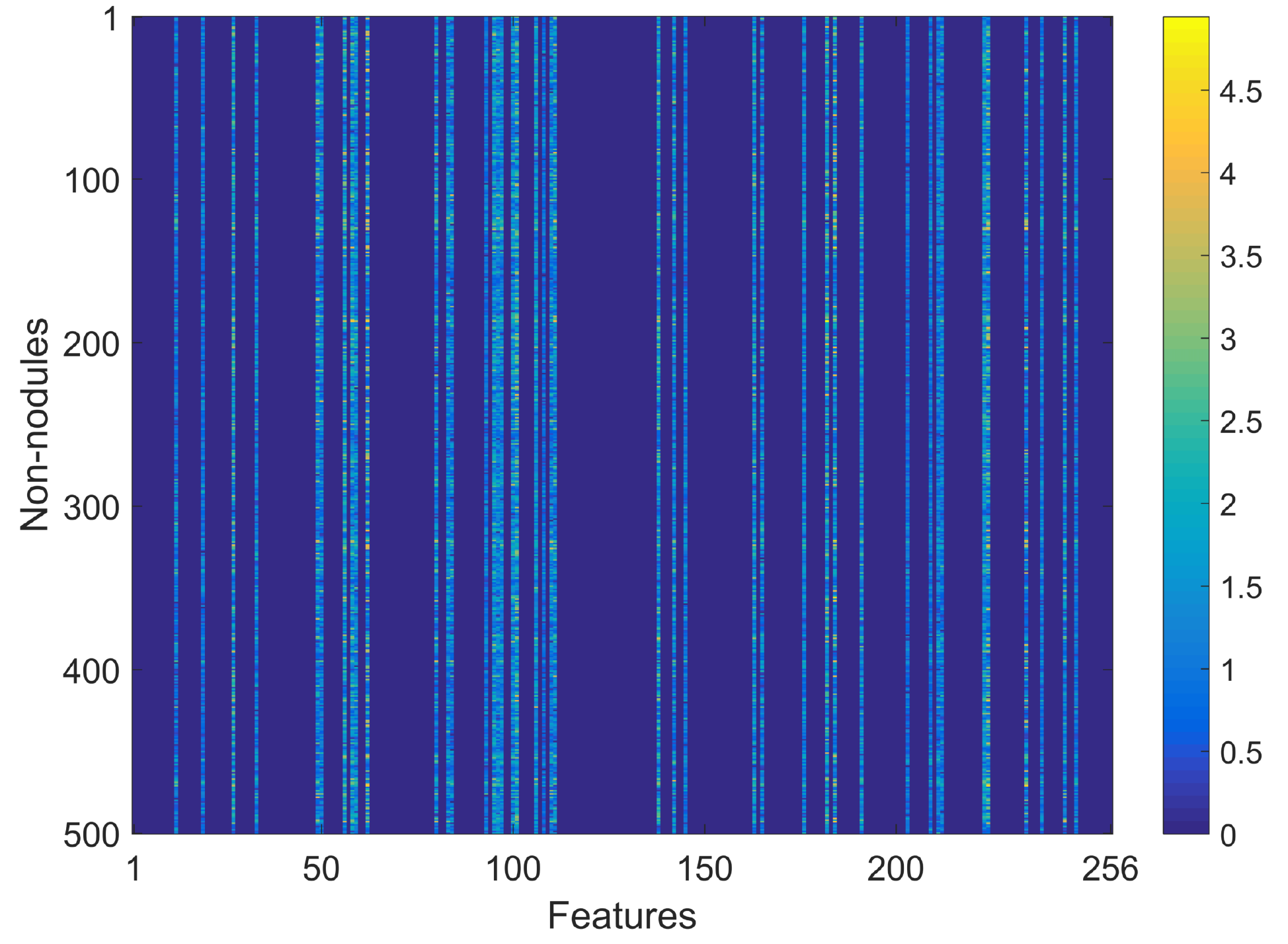}} \\
\caption{\label{fig4}
Feature matrix of non-nodule samples.}
\end{figure}

\begin{figure}[t!]
\centering{\includegraphics[width=1.0\linewidth]{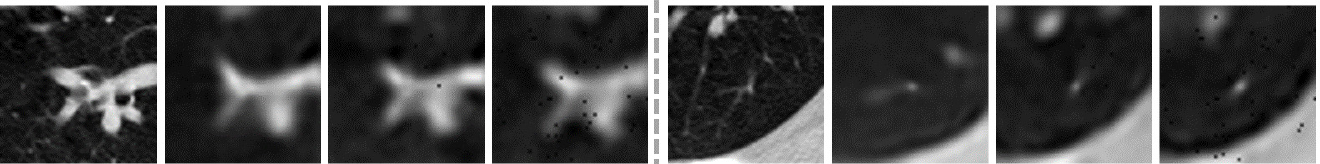}} \\
\caption{\label{fig5}
Reconstruction results depend on the feature dimension (input, 64, 256, and 512 from the left to right).}
\end{figure}
\begin{figure}[t!]
\centering{\includegraphics[width=0.8\linewidth]{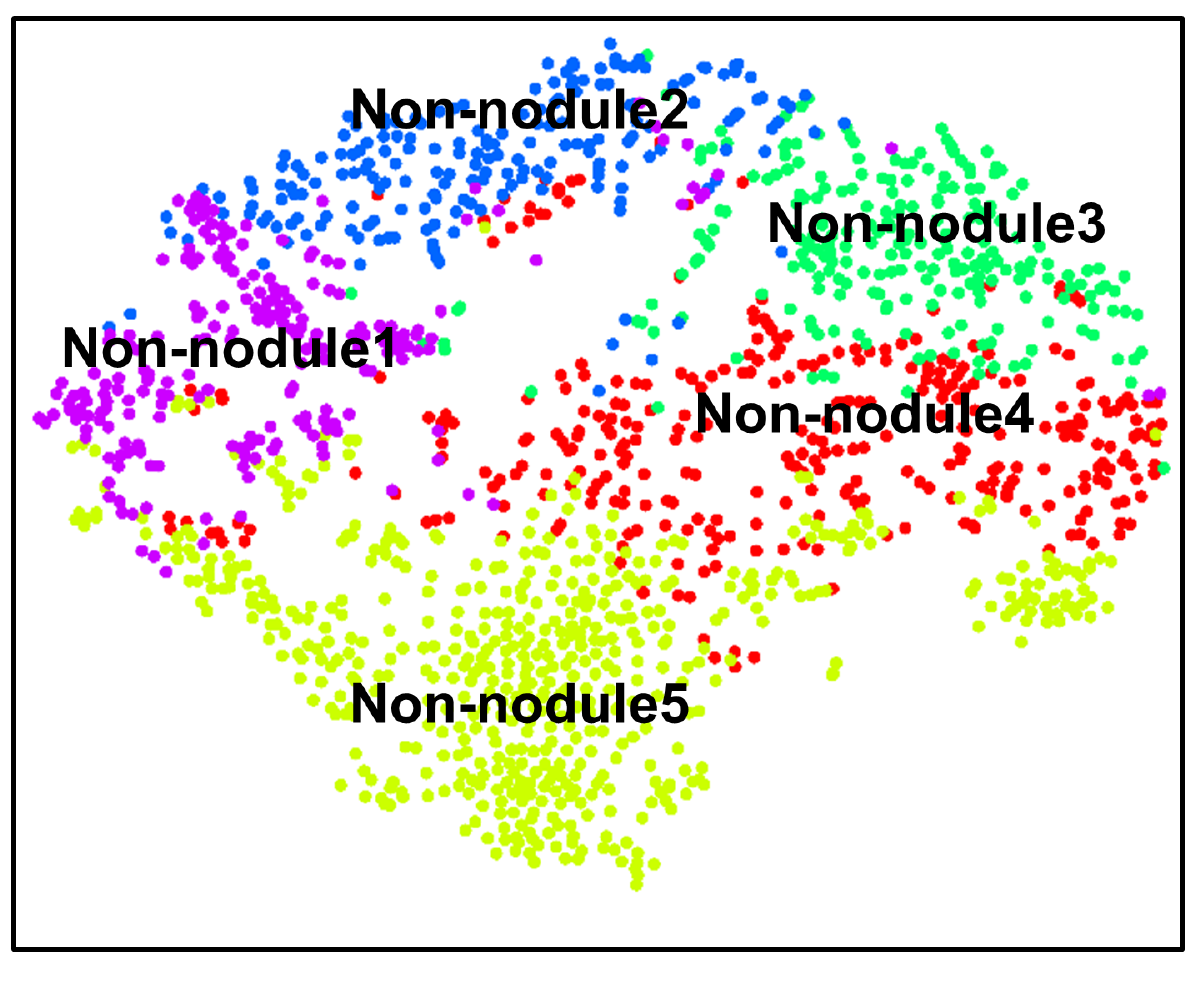}} \\
\caption{\label{fig6}
Visualization of clustered non-nodule samples (1,500 Non-nodules are randomly selected).}
\end{figure}

\subsubsection{Clustering}
We categorize non-nodules into $K$ groups according to the feature vector $\bm{h}^4$, where $K$ is the number of categories and 2D CNNs.
We first reduce the dimension of the feature vector by eliminating zeros that equally occurs on all non-nodules at the same feature dimension.
As can be seen in Fig. 4, most features have zero values on all non-nodules. Because these features cannot be used to categorize non-nodules, we simply eliminate them to only keep meaningful features.
The remaining number of features is 72.34 (28.26\%) on average over the number of folds in cross-validation.
Note that we have also extracted the feature vectors with 512 and 64 dimensions by changing the fourth layer of the autoencoder.
Feature matrices of feature vectors with 512 and 64 dimensions also have sparse structures similar to Fig. 4.
Therefore, we have chosen 256 dimensions that yield the lowest loss of the autoencoder among three types of dimensions.
Although we used 256 dimensions based on the loss, 512 and 64 dimensions also successfully reconstruct input non-nodules as shown in Fig. 5.
Based on the remained features, we perform k-means clustering with $K$ clusters.
In Fig. 6, we show clustering results with $K=5$ on 1,500 non-nodules randomly selected.
For visualization, we use t-SNE \cite{vanJMLR2008} that reduces multiple dimensions of features to two dimensions.

\subsection{Nodule probability prediction}
For predicting nodule probability of candidates, we introduce the ensemble of 2D CNNs.
Employing a 2D CNN with single-view 2D patches has many advantages in terms of memory and computational efficiency.
However, only using a 2D single view might yield unstable predictions due to less information compared to the 3D volumetric data.
To alleviate this problem, we extend the capability of networks by training each CNN with the same nodules and a different type of non-nodules obtained from non-nodule categorization (see Fig. 2(b)).
From this training strategy, CNNs can more easily learn meaningful features from the small number of similar non-nodules than from the large number of diverse non-nodules.
Furthermore, the ensemble of these 2D CNNs reduces a large number of false positives and increases prediction performance by reducing the bias of a single network \cite{rubinJTI2015}.

\begin{figure}[t!]
\centering{\includegraphics[width=1\linewidth]{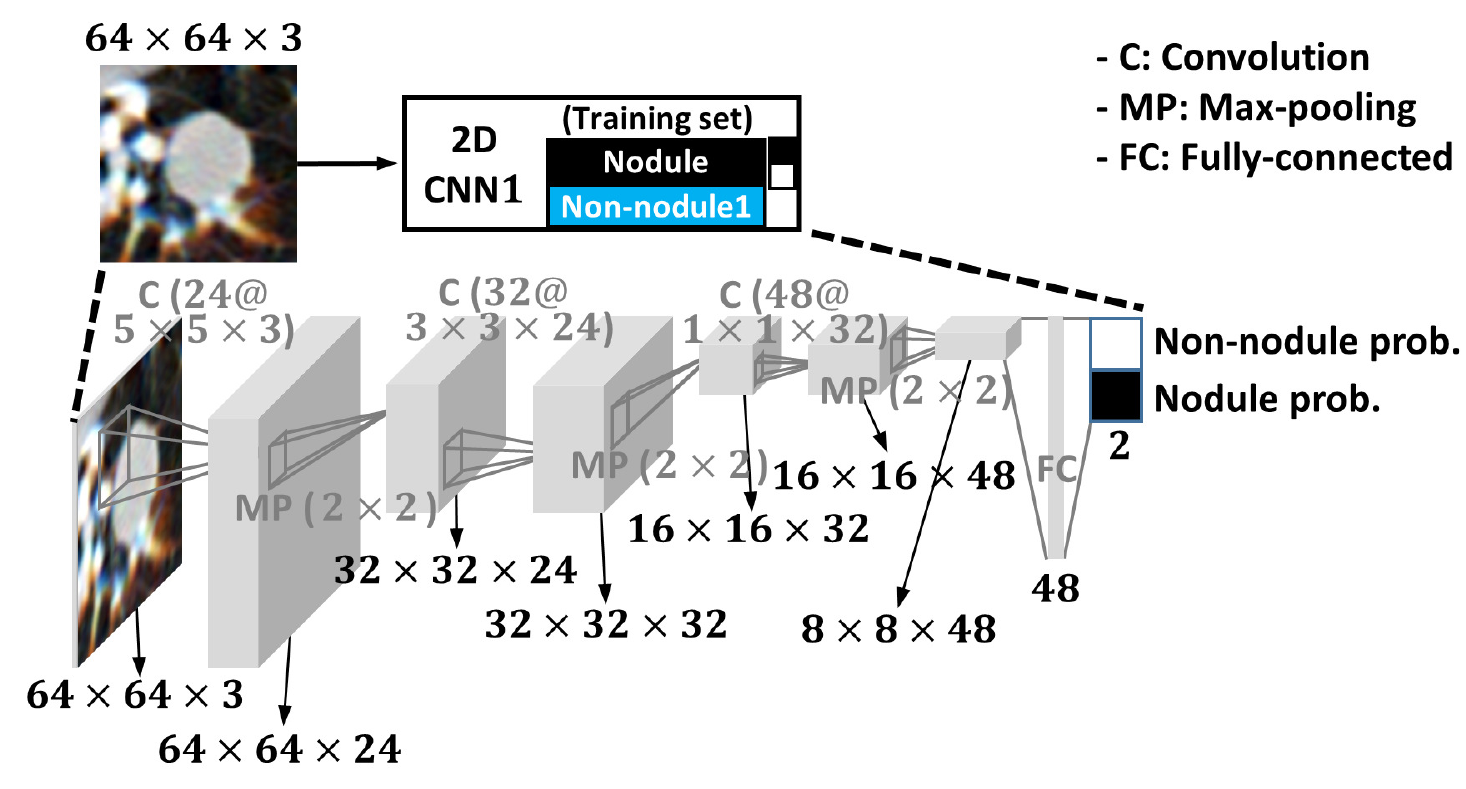}} \\
\caption{\label{fig7}
The architecture of our 2D CNN.}
\end{figure}

The architecture of the 2D CNN is illustrated in Fig. 7.
Unlike recent deep architectures used in computer vision, previous works\cite{ciompiMIA2015,setioTMI2016,douTBME2016,tajbakhshPR2017} for pulmonary nodule detection employ relatively shallow  architectures similar to ours. We have employed several architectures in terms of the number of layers.
However, the deeper architectures provide severe overfitting due to the imbalance between the number of nodules and non-nodules (we will present this technical issue in Section 5).
Additionally, deeper architectures make no significant difference in performance improvement of lung nodule detection \cite{tajbakhshPR2017}.
Note that 2D CNNs have the same architecture, but they are trained with the same nodules and different types of non-nodules.
The 2D CNN consists of three convolutional, three max-pooling, and two fully-connected layers.
The convolutional layers are formulated as follows:
\begin{eqnarray}
\bm{h}^l_i = \bm{\sigma}(\bm{b}^l_i + \sum_j (\bm{W}^l_{ji}*\bm{h}^{l-1}_j)),
\end{eqnarray}
\begin{table*}[t!]
\footnotesize
\caption{The number of nodules and non-nodules in training, validation, and test sets.}
\label{tab1}
\begin{minipage}{0.355\linewidth}
\begin{tabular}{C{1.3cm}|C{2.3cm}|C{1.8cm}}\hline 
\bf{Training}	& \bf{Nodule (Aug.)}	& \bf{Non-nodule} \\ \hline\hline
Fold 0		& 1,055	(51,695)		& 433,436 \\
Fold 1		& 1,113	(54,537)		& 431,846 \\
Fold 2		& 1,071	(52,479)		& 428,527 \\
Fold 3		& 1,055	(51,695)		& 433,598 \\
Fold 4		& 1,069	(52,381)		& 443,903 \\
Fold 5		& 1,074	(52,626)		& 433,141 \\
Fold 6		& 1,075	(52,675)		& 434,885 \\
Fold 7		& 1,077	(52,773)		& 432,494 \\ \hline

\end{tabular}
\end{minipage}
\hspace{.2cm}
\begin{minipage}{0.3\linewidth}
\begin{tabular}{C{1.6cm}|C{1.cm}|C{1.8cm}}\hline 
\bf{Validation}	& \bf{Nodule}	& \bf{Non-nodule} \\ \hline\hline
Fold 0	& 125		& 46,917			\\
Fold 1	& 97			& 49,621			\\
Fold 2	& 118		& 48,008			\\
Fold 3	& 124		& 45,351			\\
Fold 4	& 100		& 44,568			\\
Fold 5	& 100		& 45,769			\\
Fold 6	& 100		& 48,837			\\
Fold 7	& 104		& 47,097			\\ \hline
\end{tabular}
\end{minipage}
\hspace{.2cm}
\begin{minipage}{0.28\linewidth}
\begin{tabular}{C{1.2cm}|C{1.3cm}|C{1.8cm}}\hline 
\bf{Test}	& \bf{Nodule}	& \bf{Non-nodule} \\ \hline\hline
Fold 0	& 171		& 69,361			\\
Fold 1	& 141		& 68,247			\\
Fold 2	& 162		& 73,179			\\
Fold 3	& 172		& 70,765			\\
Fold 4	& 182		& 61,243			\\
Fold 5	& 177		& 70,804			\\
Fold 6	& 176		& 65,992			\\
Fold 7	& 170		& 70,123			\\ \hline
\end{tabular}
\end{minipage}
\end{table*}

where $\bm{h}^{l-1}_j$ and $\bm{h}^l_i$ denote the $j$th 2D feature in the $(l-1)$th layer and the $i$th 2D feature in the $l$th layer, respectively.
$\bm{W}^l_{ji}$ and ${b}^l_i$ are the 2D convolutional kernel and the bias term, respectively.
$\bm{\sigma}(\cdot)$ is the activation function defined as ReLU.
$*$ denotes the convolution operation.
We gradually decrease the convolutional kernel size while the feature size decreases.
We also apply zero padding to guarantee that the input and output have the same spatial size.
The max-pooling layer acquires the maximum values in a $2 \times 2$ window that slides features in a non-overlapping manner.
This progressively reduces the spatial size of features, which decreases the number of parameters.
Additionally, the max-pooling layer can control the overfitting for the generality of a network. After three convolutional layers and three consecutive max-pooling layers, we connect a fully-connected layer with 48 output neurons.
The fully-connected layer is formulated the same as the layer in an autoencoder (see Eq. (1)).
The last layer is also fully-connected with the previous layer, but we apply the softmax regression to the output to predict probabilities for each class $c$ as follows:
\begin{eqnarray}
p_c(\bm{h}^L) = \frac{e^{h^L_c}}{e^{h^L_0}+e^{h^L_1}},
\end{eqnarray}
where $\bm{h}^L$ and $h^L_c$ are the output vector of the last layer and the $c$th element of it, respectively.
For the class $c$, 0 denotes a non-nodule and 1 denotes a nodule.

Let $\{(\bm{X}_{\tiny{\text{CNN}}}^{(1)},y^{(1)}), ... , (\bm{X}_{\tiny{\text{CNN}}}^{(N)},y^{(N)})\}$ denote a set of $N$ labeled training samples.
$\bm{X}_{\tiny{\text{CNN}}}^{(n)}$ is an input patch of $64 \times 64 \times 3$ size on axial view.
$y^{(n)}$ is the label of the corresponding patch.
The input patch consists of three 64$\times$64 patches, including a patch where a candidate nodule is located at the center and two adjacent patches.
Given the input patch, more depth would provide more information. For training all parameters $\theta_{\tiny{\text{CNN}}}$ of the 2D CNN, we optimize the loss function formulated as
\begin{eqnarray}
\begin{aligned}
& \mathcal{L}(\theta_{\tiny{\text{CNN}}}) = -\frac{1}{N}\bigg\{ \sum^N_{n=1} \sum^{C-1}_{c=0}\textbf{1}\{y^{(n)}=c\} \\
& \phantom{\mathcal{L}(\theta_{\tiny{\text{CNN}}}) =} \times \text{log}P(\hat{y}^{(n)}=c|\bm{X}_{\tiny{\text{CNN}}}^{(n)};\theta_{\tiny{\text{CNN}}}) \bigg\},
\end{aligned}
\end{eqnarray}
where $\bm{1}\{\cdot\}$ is the indicator operation and $\hat{y}^{(n)}$ is the predicted label.
$P(\hat{y}^{(n)}=c|\bm{X}_{\tiny{\text{CNN}}}^{(n)};\theta_{\tiny{\text{CNN}}})$, which equals to $p_c(\bm{h}^L)$, is the estimated probability of the input $\bm{X}_{\tiny{\text{CNN}}}^{(n)}$ for the class $c$.
For initial parameters, the convolution kernels and the weights of fully-connected layers were randomly initialized from Gaussian distribution $\mathcal{N}(0, 0.05^2)$ and $\mathcal{N}(0, 0.04^2)$, respectively.
The biases for all layers were initialized as zeros. Every parameter was updated with Adam optimization same as the autoencoder.
We initialized the learning rate as 0.001 with 4\% decay on every 500 iterations.
For fully-connected layers, we utilized the dropout \cite{hintonarXiv2012} with rate 0.5 that improves the generality of a network.
All hyper-parameters for training were experimentally determined.

After training 2D CNNs, we can obtain $K$ pairs of a nodule and a non-nodule probability for a testing candidate patch $\bm{X}_{\tiny{\text{CNN}}}^{(n)}$.
We simply average these $K$ pairs of the probabilities to estimate the final prediction probability $P_{F}(\hat{y}^{(n)}=c|\bm{X}_{\tiny{\text{CNN}}}^{(n)};\theta_{\tiny{\text{CNN}}})$ at the fusion stage:
\begin{eqnarray}
\begin{aligned}
& P_{F}(\hat{y}^{(n)}=c|\bm{X}_{\tiny{\text{CNN}}}^{(n)};\theta_{\tiny{\text{CNN}}}) \\
& = \frac{1}{K}\sum^K_{k=1} P_{k}(\hat{y}^{(n)}=c|\bm{X}_{\tiny{\text{CNN}}}^{(n)};\theta_{\tiny{\text{CNN}_k}}),\!\!\!
\end{aligned}
\end{eqnarray}
where $k$ is the index of 2D CNNs.

\section{Experiments}
For implementing our proposed framework, we used TensorFlow \cite{abadiarXiv2016} with a GPU of NVIDIA GeForce GTX 1070.
We performed eight-fold cross-validation based on the number of CT scans of the LUNA16 database.
We divided all scans to construct a test set, a training set, and a validation set as illustrated in Fig. 8.
First, we randomly divided 888 CT scans into eight subsets that contain the equal number of CT scans.
Among these eight subsets, we tested a single subset with 111 CT scans.
\begin{figure}[t!]
\footnotesize
\begin{minipage}{0.45\linewidth}
\centering{\includegraphics[width=0.95\linewidth]{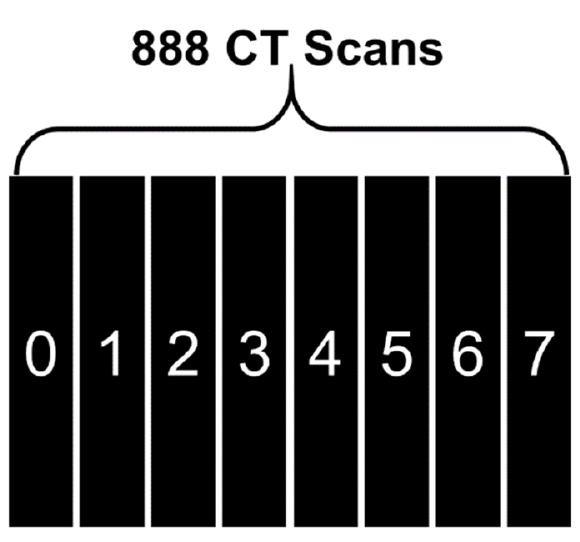}}
\centering{(a)}
\end{minipage}
\begin{minipage}{0.45\linewidth}
\centering{\includegraphics[width=0.95\linewidth]{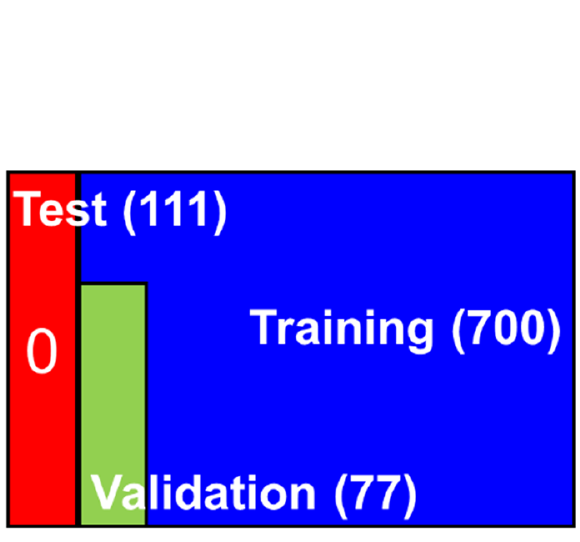}}
\centering{(b)}
\end{minipage}
\caption{\label{fig8}
Dataset construction. We (a) randomly divide 888 CT scans into eight subsets to perform the eight-fold cross-validation. Among eight subsets, we (b) construct a test set with 111 CT scans (i.e., a single subset), a training set with 700 CT scans, and a validation set with 77 CT scans.}
\end{figure}
A training set and a validation set consist of 90\% and 10\% of the rest 777 CT scans, respectively.
We used a validation set to check the time to stop training 2D CNNs, which prevents the overfitting.
Also, it can help to estimate the accuracy of a test set through an accuracy of a validation set.
We describe the number of nodules and non-nodules in training, validation, and test sets in Table 1.
The imbalance problem between the number of nodules and non-nodules is severe in training sets.
To solve this problem, we augmented nodules by 49 times through translation, rotation, and flipping.
Also, we normalized intensities to $[0, 1]$ following clipping the intensities into the interval (-1000, 400 Hounsfield Unit) as pre-processing.

\begin{figure}[t!]
\footnotesize
\centering{(a)}
\centering{\includegraphics[width=0.9\linewidth]{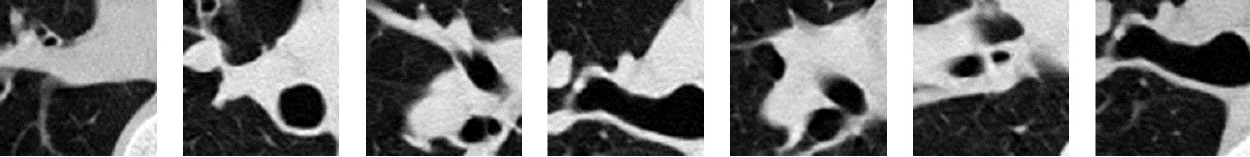}}\\
\centering{(b)}
\centering{\includegraphics[width=0.9\linewidth]{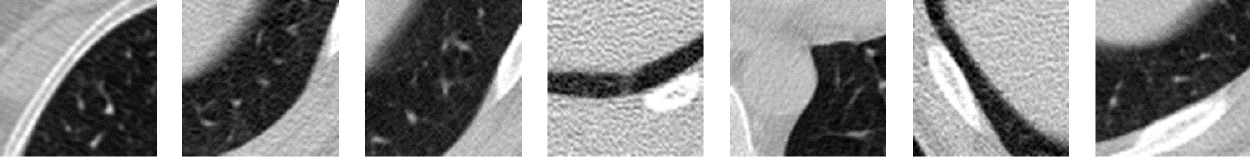}}\\
\centering{(c)}
\centering{\includegraphics[width=0.9\linewidth]{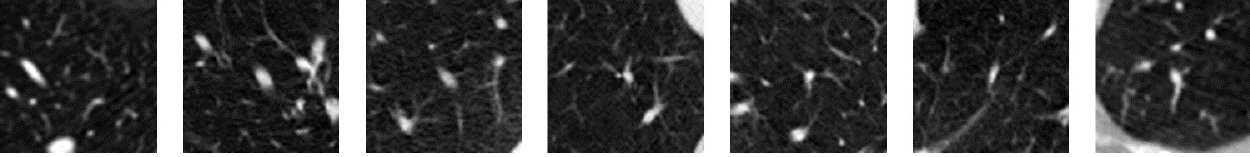}}\\
\centering{(d)}
\centering{\includegraphics[width=0.9\linewidth]{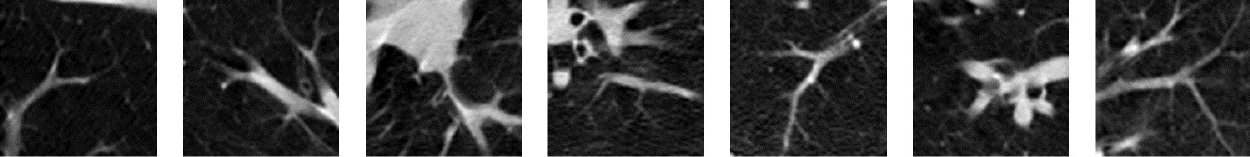}}\\
\centering{(e)}
\centering{\includegraphics[width=0.9\linewidth]{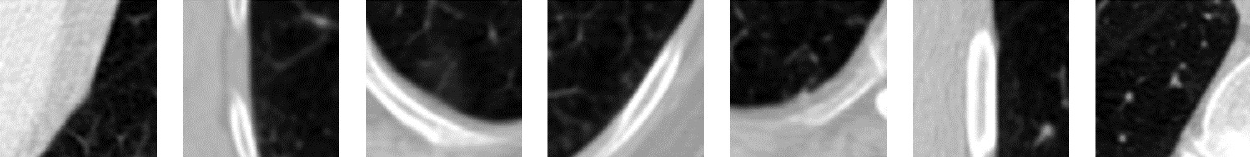}}\\
\caption{\label{fig9}
Categorized non-nodules that account for a majority of each category on the fold 1. (a) Non-nodule 1. (b) Non-nodule 2. (c) Non-nodule 3. (d) Non-nodule 4. (e) Non-nodule 5.}
\end{figure}

\subsection{Non-nodule categorization}
We show categorized non-nodules accounting for a majority of each group in Fig. 9.
We set $K=5$.
Note that non-nodule samples for non-nodule categorization are nodule candidates extracted from candidate detection.
We do not randomly select non-nodule samples as everything else but nodules in a scan.
The random selection might provide poor performance in modeling data since the variance in non-nodule samples would be very large.
As can be seen, each group involves visually similar non-nodules.
For example, non-nodule 1 commonly contains arbitrary and thick structures.
In non-nodule 2 and 5, chest walls are heavily included.
In case of non-nodule 5, lung regions are more homogeneous compared to non-nodule 2.
Non-nodule 3 and 4 generally contain vessel branches.
However, vessels in non-nodule 4 are clearer and thicker than ones in non-nodule 3.
In Fig. 10, we also show several non-nodules including different patterns in the same category.
Note here that it is difficult to categorize a wide variety of non-nodules with the limited number of groups.
This problem would be alleviated by increasing the number of categories, while more computational costs are required.
Also, a denoising autoencoder \cite{vincentJMLR2010} can be utilized instead of an autoencoder to extract features of non-nodules.
The denoising autoencoder is trained with a corrupted input to reconstruct a clean version of it.
From the reconstruction process, the denoising autoencoder would learn more general and higher representations \cite{vincentJMLR2010}.
In Fig. 11, reconstruction results of the autoencoder and denoising autoencoder are presented for visual comparison.
We will show performance comparisons between utilizing a denoising autoencoder and an autoencoder for feature extraction in Section 5.2.

\begin{figure}[t!]
\centering{\includegraphics[width=0.95\linewidth]{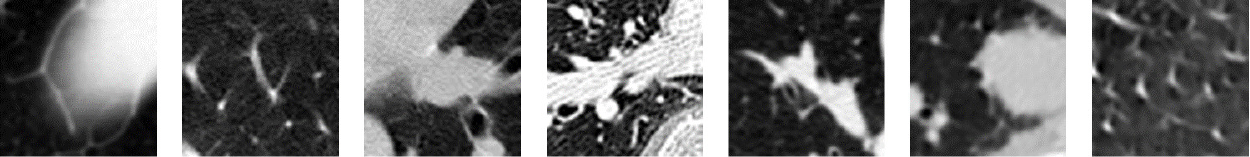}}
\caption{\label{fig10}
A wide range of non-nodules in the same category.}
\end{figure}
\begin{figure}[t!]
\footnotesize
\centering{(a)}
\centering{\includegraphics[width=0.9\linewidth]{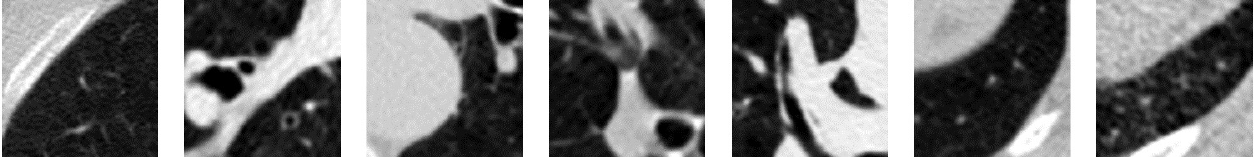}}\\
\centering{(b)}
\centering{\includegraphics[width=0.9\linewidth]{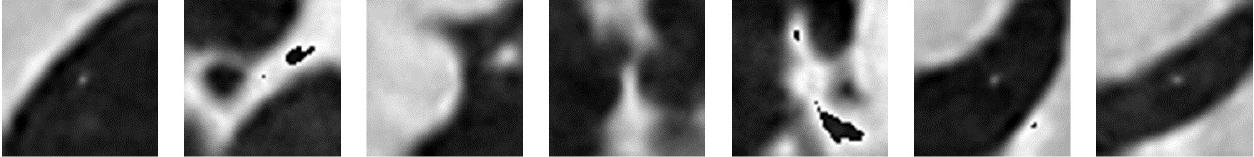}}\\
\centering{(c)}
\centering{\includegraphics[width=0.9\linewidth]{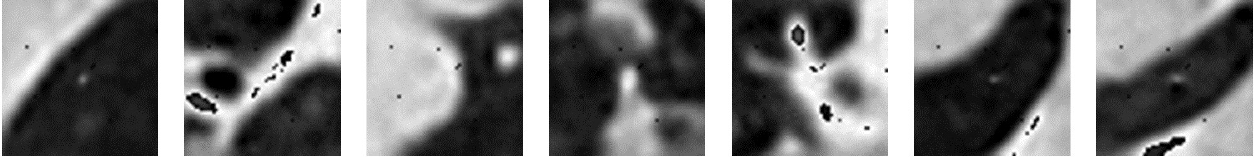}}\\
\caption{\label{fig11}
Reconstruction results. (a) Reference non-nodules. (b) Reconstructed non-nodules from an autoencoder. (c) Reconstructed non-nodules from a denoising autoencoder.}
\end{figure}

\subsection{Nodule probability prediction}
We assessed the effectiveness of our non-nodule categorization and several architectures.
We experimented on five training sets including differently categorized non-nodules as shown in Fig. 12.
Note that we set $K=5$ to consider the balance between the number of nodules and non-nodules in each training set.
First, we used our non-nodule categorization based on an autoencoder.
Second, we categorized non-nodules by using non-nodule categorization based on a denoising autoencoder.
Third, we randomly divided non-nodules in equal number to construct different non-nodule sets, which replaces our non-nodule categorization.
Through comparison between architectures trained using the randomly divided non-nodules and the categorized non-nodules, we validated the effectiveness of our non-nodule categorization.
Next, we trained all 2D CNNs by using same nodules and same non-nodules (i.e., all training data are used to train each 2D CNN).
This can be a baseline to confirm the advantage of training networks using different types of non-nodules.
Last, we utilized a single 2D CNN trained with all nodules and non-nodules as a baseline for false positive reduction utilizing a 2D single view.
Throughout the rest of the paper, we refer to the first, second, third, fourth, and fifth architectures as \textbf{AE}, \textbf{DAE}, \textbf{R}, \textbf{A}, and \textbf{S}, respectively.

\begin{figure}[t!]
\centering{\includegraphics[width=.49\linewidth]{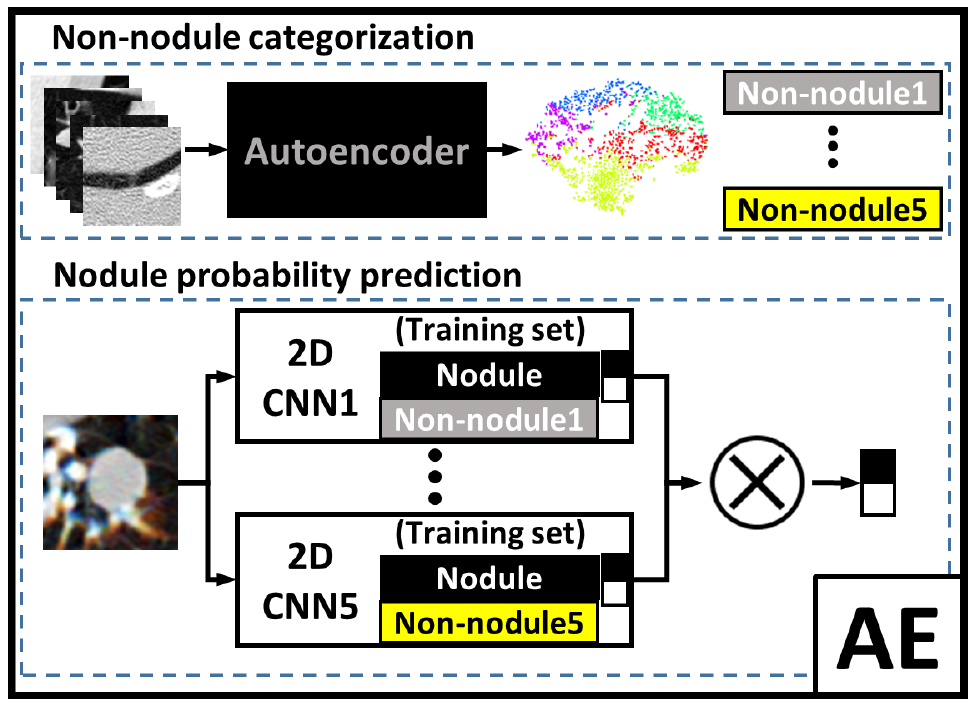}}\\
\centering{\includegraphics[width=.49\linewidth]{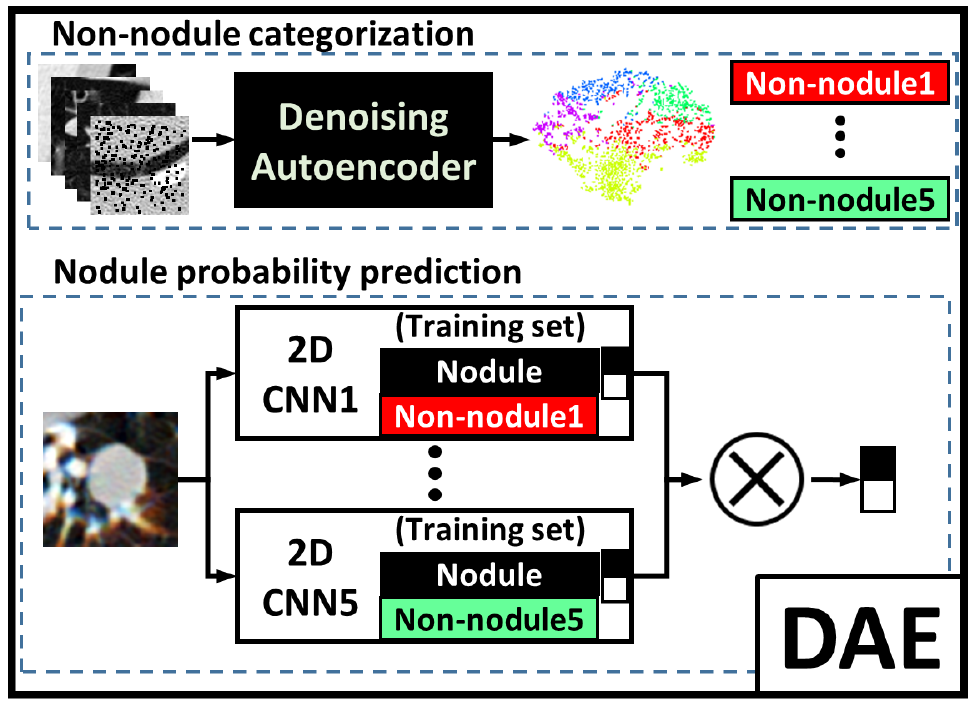}}
\centering{\includegraphics[width=.49\linewidth]{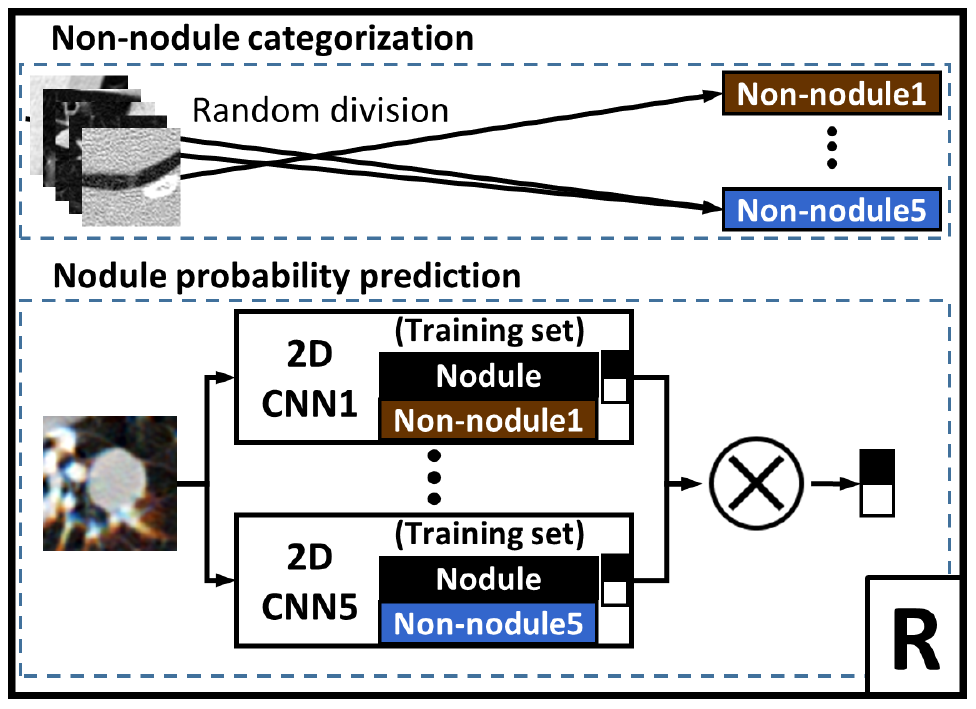}}
\centering{\includegraphics[width=.49\linewidth]{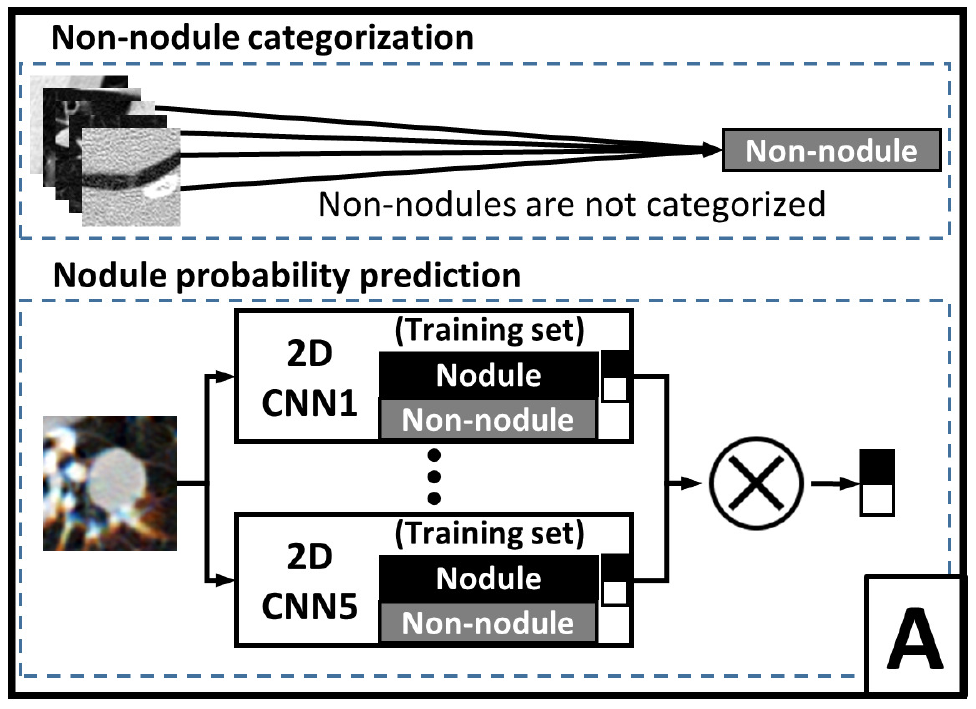}}
\centering{\includegraphics[width=.49\linewidth]{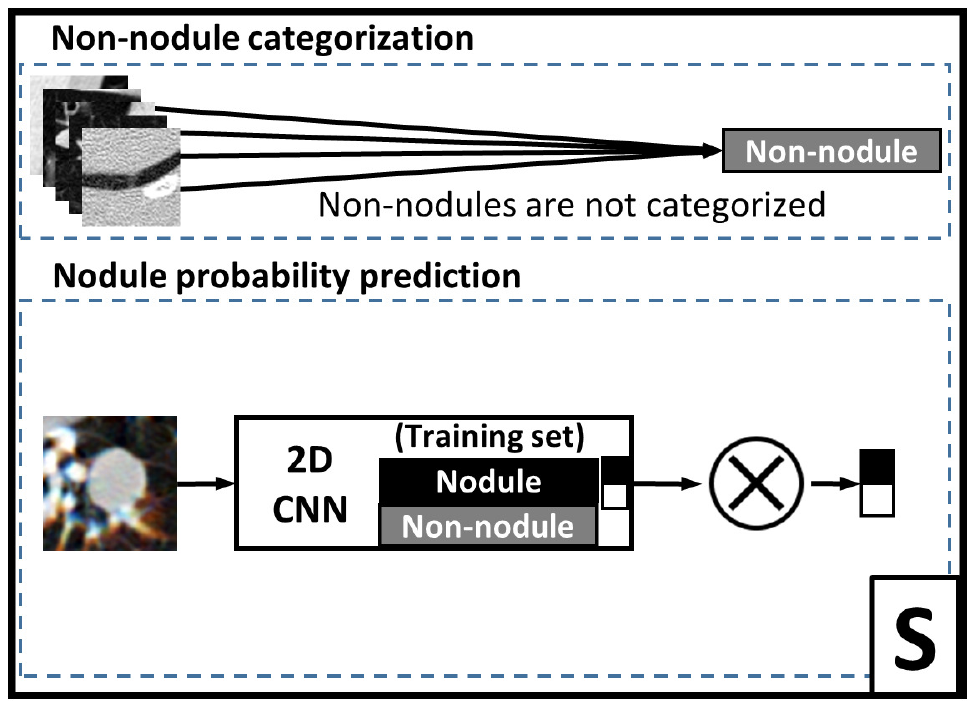}}
\caption{\label{fig12}
Five frameworks according to differently constructed non-nodule training sets (color difference of non-nodule training sets indicates that non-nodules in each set are different). Same nodule sets are utilized for all CNNs. \textbf{AE}: CNNs are trained with non-nodule training sets categorized by our non-nodule categorization with an autoencoder. \textbf{DAE}: CNNs are trained with non-nodule training sets categorized by our non-nodule categorization with a denoising autoencoder. \textbf{R}: CNNs are trained with non-nodule training sets randomly divided in equal number. \textbf{A}: every CNN is trained with all non-nodules. \textbf{S}: a single CNN is trained with all non-nodules.}
\end{figure}

\begin{figure}[t!]
\centering{\includegraphics[width=0.9\linewidth]{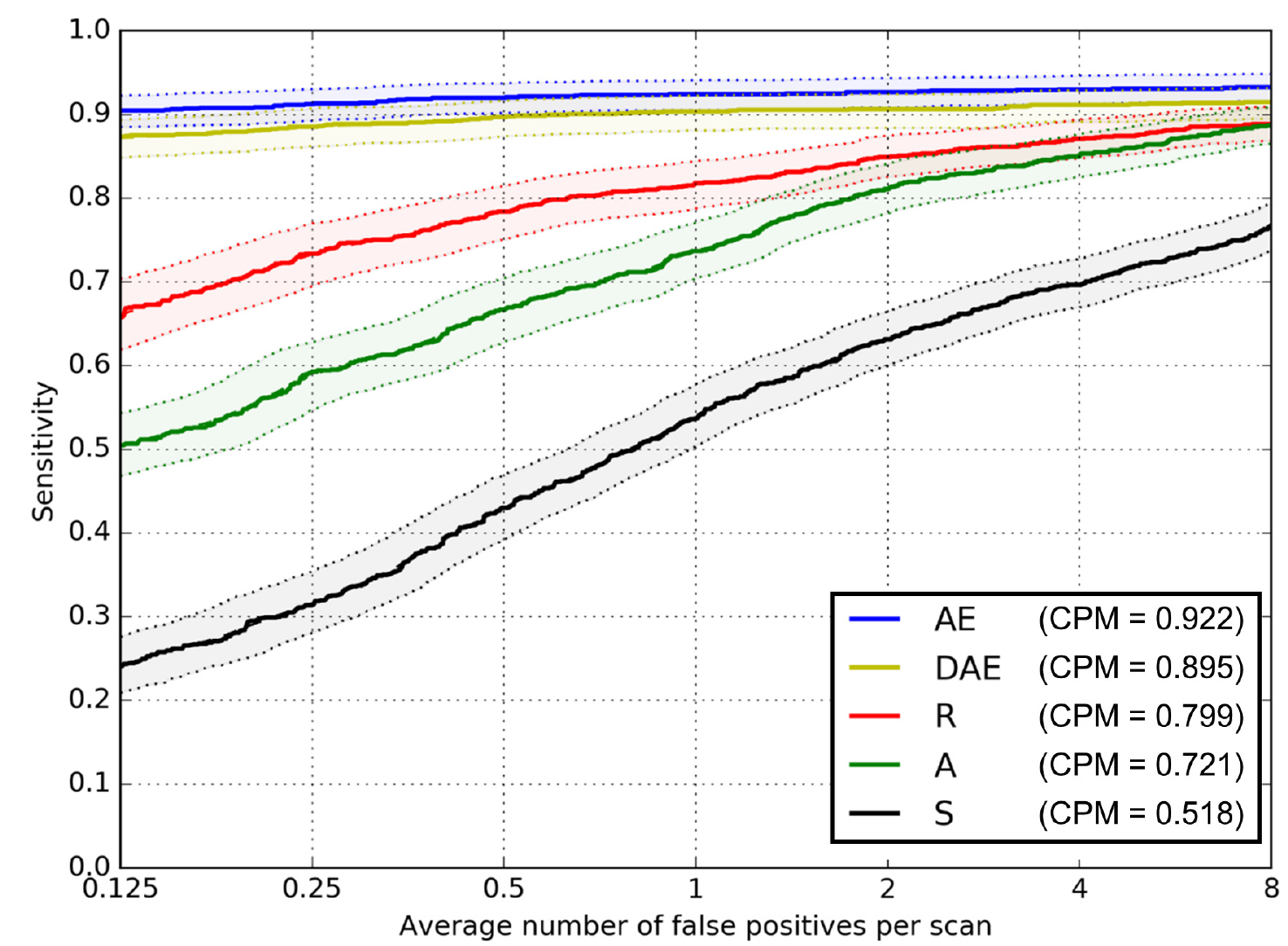}}\\
\caption{\label{fig13}
FROC curves of five frameworks with different non-nodule training sets (95\% confidence interval is indicated as a dash line).}
\end{figure}
\begin{figure}[t!]
\footnotesize
\centering{(a)}
\centering{\includegraphics[width=0.90\linewidth]{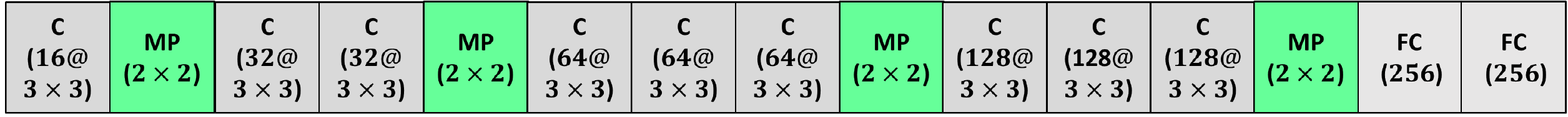}}\\
\centering{\includegraphics[width=0.99\linewidth]{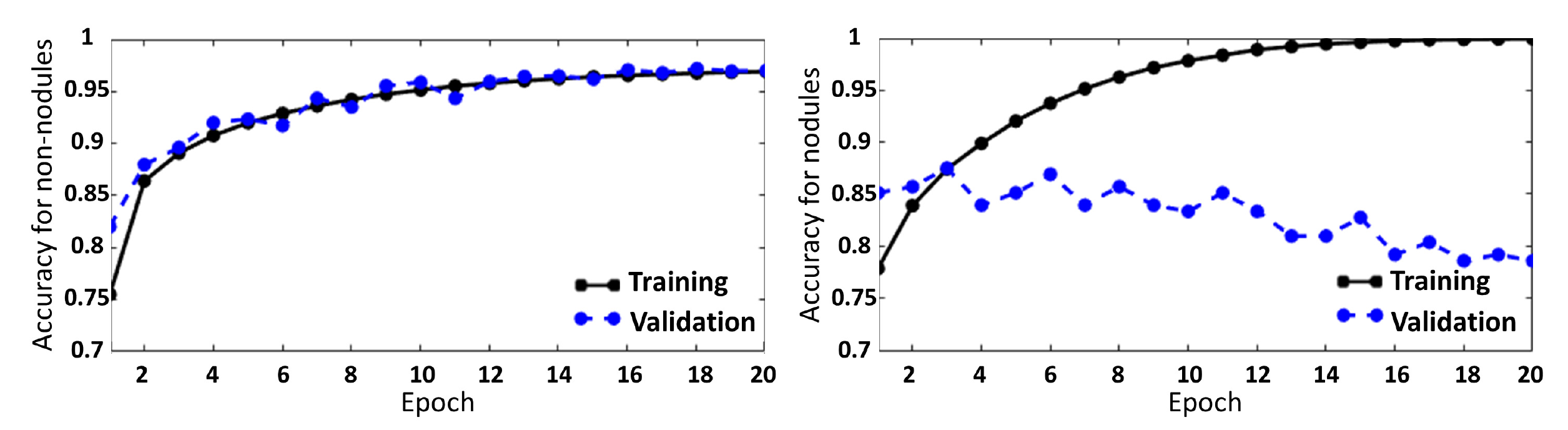}}\\
\centering{(b)}\\
\caption{\label{fig14}
Overfitting results of a deeper architecture. (a) Description of a deeper architecture. (b) Accuracy of non-nodules (left) and nodules (right)}
\end{figure}
For evaluating the performance, we plot a free-response receiver operating characteristic (FROC) curve with the 95\% interval using bootstrapping \cite{efron1994} in Fig. 13.
We used the public code released by the LUNA16 challenge to draw FROC curves.
Test, training, and validation sets are constructed as shown in Fig. 8 and Table 1.
The x-axis and the y-axis of FROC curves are defined as the sensitivity on nodules and average numbers of false positives per scan, respectively.
In Fig. 13, the FROC curves of five different frameworks are presented.
The performance of \textbf{S} is much lower than others because the prediction of a single network would be easily biased.
Moreover, a single 2D CNN would have difficulty to learn representative features of non-nodules with large appearance variations.
Note that we also employed a deeper architecture.
However, the deeper architecture provides overfitting results especially on nodules (see Fig. 14).
As the comparison between the performances of \textbf{A} and \textbf{S}, we can notice that an ensemble scheme increases the performance by alleviating the bias of a single network.
The training each network with different non-nodules also efficiently improves the performance of false positive reduction (see the gap between the curves of \textbf{R} and \textbf{A}).
Last but not least, \textbf{AE} and \textbf{DAE} using non-nodule sets categorized by our non-nodule categorization achieves the best and the second best performance, respectively.
From these results, it is demonstrated that our non-nodule categorization can effectively assist false positive reduction in pulmonary nodules detection regarding network training.

\begin{figure}[t!]
\centering{\includegraphics[width=0.9\linewidth]{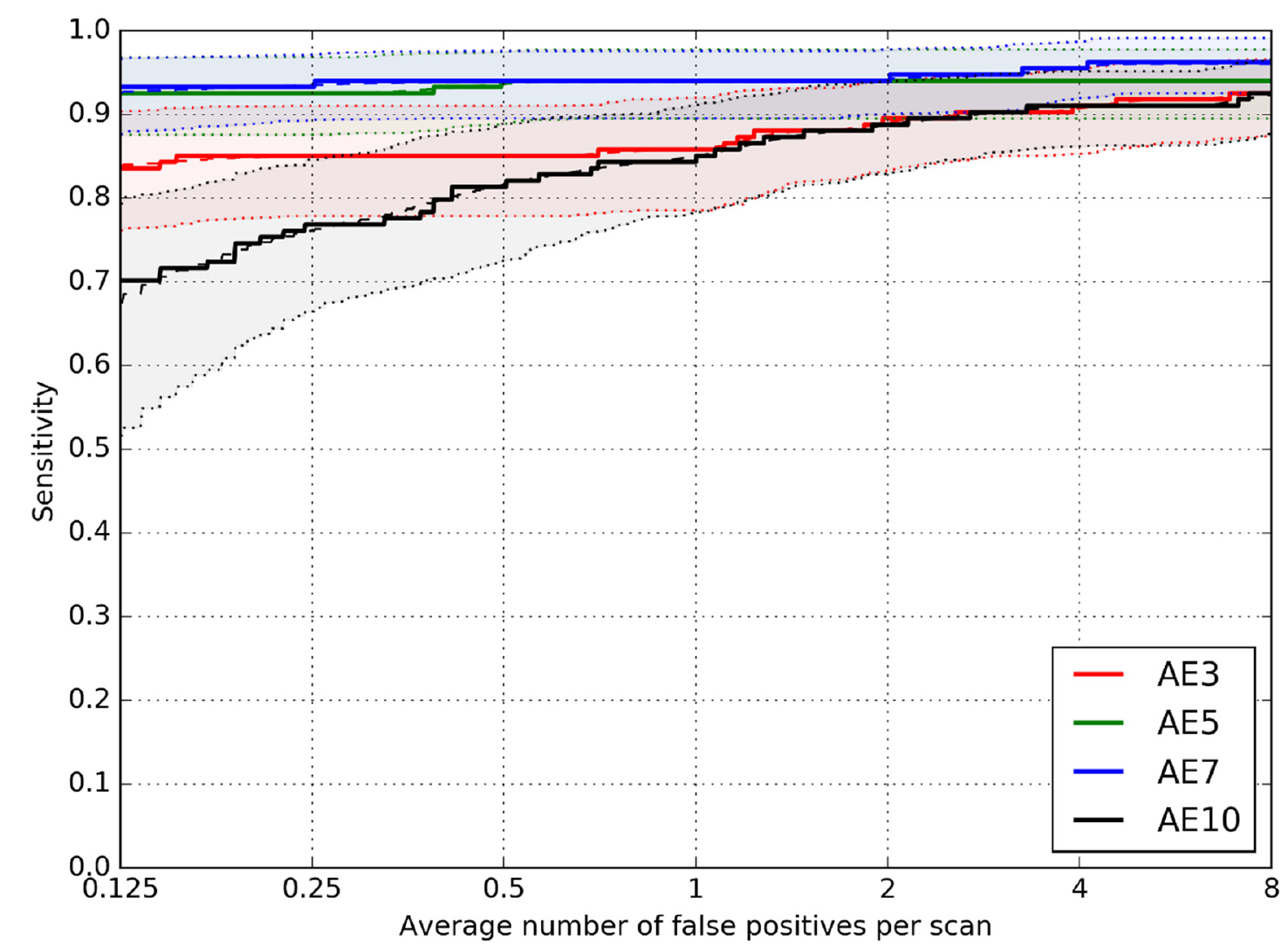}}\\
\caption{\label{fig15}
FROC curves of the different number of non-nodule categories on the fold 0 dataset (the number in the legend indicates the number of non-nodule categories $K$).}
\end{figure}

We additionally compared performances by increasing $K$ to find the proper number of non-nodule categories in Fig. 15.
Among four different $K$ settings, \textbf{AE10} (i.e., $K=10$) yields the worst performance due to the imbalance between the number of nodule and non-nodule samples for training.
The larger $K$ value would more specifically categorize the types of non-nodules.
However, we observed that the gap between the number of samples in each category also increases.
In other words, with increasing $K$, a dominant category including the large number of samples is obtained, but the rest of categories consist of the small number of samples.
This causes the significant imbalance problem in training.
\textbf{AE3} shows the better performance than \textbf{AE10} but it is still low.
We believe that it is difficult for various non-nodule samples to be categorized with a small $K$ value.
We achieved the best performance with \textbf{AE7} and also the similar performance with \textbf{AE5}.
According to these results, it is important to select the number of categories $K$ while considering the balance of training data.

\begin{table*}[t]
\footnotesize
\centering
\caption{Architecture description of teams participated in false positive reduction track in the LUNA16 challenge.}
\label{tab1}
\begin{tabular}{C{3.5cm}|C{2.5cm}|C{1.4cm}|C{1.7cm}|C{1.7cm}|C{2.cm}|C{2.cm}}\hline 
\bf{Team}	& \bf{Data Type} & \bf{Ensemble} & \bf{Number of Networks} & \bf{Number of Layers} & \bf{Number of Parameters} & \bf{FLOPS}\\ \hline\hline
\bf{JianPeiCAD} 					& 3D 			& N/A	  & N/A & N/A & N/A  	& N/A \\
\bf{GIVECAD} 						& 3D 			& $\times$ & N/A & N/A & N/A 	& N/A \\
\bf{realize} 						& N/A			& N/A 	  & N/A & N/A & N/A  	& N/A \\
\bf{JackFPR$\_$ma}					& 3D				& $\circ$  & N/A & N/A & N/A 	& N/A \\
\bf{CUMedVis} 					& 3D				& $\circ$  & 3   & 5   & 1,422K & 5,961M\\
\bf{DIAG$\_$CONVNET} 				& 2D (Nine views)	& $\circ$  & 9   & 5   & 442K	& 2,147M\\
\bf{Ours (AE5)} 					& 2D (Single view)	& $\circ$  & 5   & 5   & 789K 	& 1,024M\\ \hline
\end{tabular}
\\ \begin{flushright}
\vspace{-0.25cm}
*N/A: No description is available in the LUNA16 website.
\end{flushright}
\vspace{-0.75cm}
\end{table*}

\begin{table*}[t]
\footnotesize
\centering
\caption{Sensitivity according to average FPs/Scan of false positive reduction track in the LUNA16 challenge.}
\label{tab2}
\begin{tabular}{C{3.5cm}|C{1.35cm} C{1.35cm} C{1.35cm} C{1.35cm} C{1.35cm} C{1.35cm} C{1.35cm}|C{1.35cm}}\hline 
\diagbox[width=3.6cm]{\bf{Teams}}{\bf{FPs / Scan}} & \bf{0.125} & \bf{0.25} & \bf{0.5} & \bf{1} & \bf{2} & \bf{4} & \bf{8} & \bf{CPM}\\ \hline\hline
\bf{JianPeiCAD} 					& 0.743 & 0.826 & 0.896 & 0.931 & 0.940 & 0.943 & 0.944 & 0.889 \\
\bf{GIVECAD} 						& 0.735 & 0.795 & 0.841 & 0.874 & 0.903 & 0.919 & 0.933 & 0.857 \\
\bf{realize} 						& 0.731 & 0.782 & 0.834 & 0.873 & 0.901 & 0.915 & 0.926 & 0.852 \\
\bf{JackFPR$\_$ma}					& 0.713 & 0.790 & 0.839 & 0.864 & 0.892 & 0.909 & 0.921 & 0.847 \\
\bf{CUMedVis} 		 			& 0.678 & 0.738 & 0.816 & 0.848 & 0.879 & 0.907 & 0.922 & 0.827 \\
\bf{DIAG$\_$CONVNET} 				& 0.692 & 0.771 & 0.809 & 0.863 & 0.895 & 0.914 & 0.923 & 0.838 \\
\bf{Ours (AE5)} 					& 0.905 & 0.913 & 0.921 & 0.925 & 0.927 & 0.931 & 0.933 & \bf{0.922} \\ \hline
\end{tabular}
\end{table*}

We performed a comparison between works of top-6 teams participated in the false positive track of the LUNA16 challenge.
We referred the LUNA16 leaderboard system for details of these teams.
Note that we and the teams that use the initially provided list of nodule candidates are indicated with an asterisk on the leaderboard system.
Table 2 describes architectures of top-6 teams and ours.
For data type, four teams used 3D patches to obtain rich information.
However, it is extremely expensive in computational cost to utilize 3D CNNs with 3D data.
By contrast, DIAG\_CONVNET \cite{setioTMI2016} and our architecture exploited 2D patches to consider computational and storage efficiency.
We also calculated the number of parameters and floating point operations per second (FLOPS) of each architecture to measure computational complexity.
Even though CUMedVis \cite{douTBME2016} utilizes the smaller number of networks than DIAG\_CONVNET and ours, CUMedVis requires the highest computational complexity with the number of parameters of 1,422K and FLOPS of 5,961M due to employing 3D CNNs.
We measured the training time and the test time for our proposed framework.
The average training time for an autoencoder and a single CNN are 2.1 hours and 1 hours, respectively.
For the k-means clustering to categorize non-nodules, it takes only 10 secs.
The test time per scan including 621 candidates on average is 0.49 secs.
In Table 3, we present sensitivities on seven false positives numbers (i.e., 0.125, 0.25, 0.5, 1, 2, 4, and 8) and a competition performance metric (CPM) score \cite{niemeijerTMI2011}.
A CPM score is computed an average sensitivity at seven false positive numbers.
As can be seen, sensitivities at small false positive numbers of other teams are low because non-nodules similar to nodules obtain high nodule probabilities.
Unlike them, the proposed framework (i.e., \textbf{AE5}) obtain relatively high sensitivities even at small false positive numbers. 
Furthermore, our framework achieved the highest CPM score even with low computational complexity.

We additionally performed a comparison between 1) Multi-MTANN which follows the similar framework to ours and 2) AE5.
We referred to the FROC curve in \cite{suzukiMP2003} for the comparison.
Note that the Multi-MTANN experimented on the test set consisting of 57 nodules and 1,726 non-nodules and our test set contains 169 nodules and 68,714 non-nodules on average.
The FROC curve of Multi-MTANN was computed based on the number of false positives per \textit{section}.
However, we used the number of false positives per \textit{scan} to compute an FROC curve of our method (AE5).
Although we evaluated performance on more challenging setting (i.e., the larger dataset and sensitivity against the number of false positives per \textit{scan}), we achieve higher performance than Multi-MTANN as shown in Fig. 16.

\begin{figure}[t!]
\centering{\includegraphics[width=0.9\linewidth]{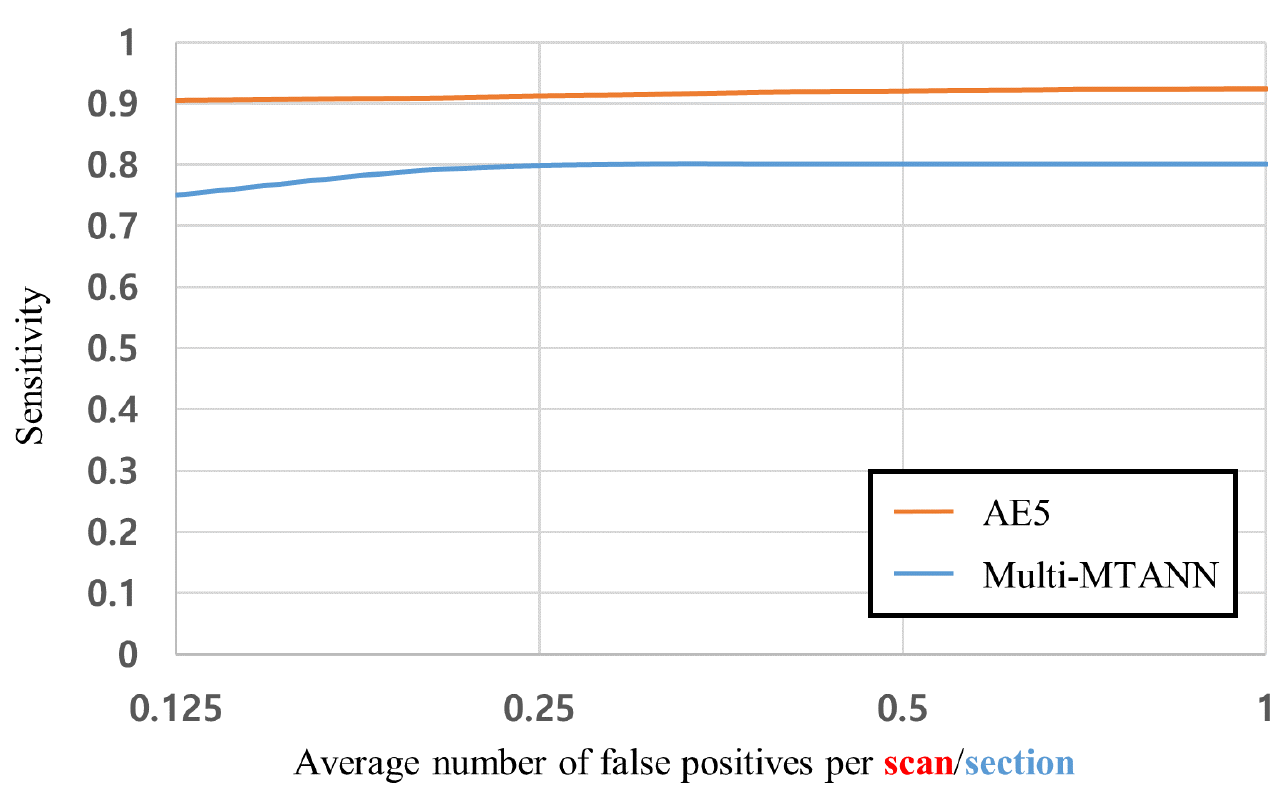}}\\
\caption{\label{fig16}
FROC curves of AE5 and Multi-MTANN.}
\end{figure}

\section{Discussion}
The proposed framework performs false positive reduction in pulmonary nodule detection via single-view 2D CNNs and fully automatic non-nodule categorization.
Most previous frameworks that achieve high performance employ 3D CNNs with 3D patches to encode richer information.
Although these 3D CNNs utilize more information, it is difficult to learn representative features without networks consisting of a large number of parameters and layers.
This is because such networks need to encode more information into representative features.
However, these complicated 3D networks are difficult to be configured due to limited computation resources.

In recent state-of-the-art studies \cite{suzukiMP2003,tajbakhshPR2017}, the use of a 2D single view has been proven to be useful in the task of false positive reduction in pulmonary nodule detection.
We believe that representative features of nodules and non-nodules can be learned from a 2D single view.
For instance, a 2D single view can still provide useful information such as round shapes of nodules and mass located at the center (see Fig. 1 (a)).
Additionally, our training scheme training 2D CNNs on different types of non-nodules efficiently extends the learning capability of networks to representative features.
Please also note that a 2D single view is a nodule candidate (rather than a usual 2D image) extracted from the candidate detection.

To confirm the effectiveness of our proposed framework, we trained our architecture on five differently constructed training sets indicated as \textbf{AE}, \textbf{DAE}, \textbf{R}, \textbf{A}, and \textbf{S}.
From the improved performance of \textbf{A} compared to \textbf{S} in Fig. 13, we notice that the prediction of a single network would be easily biased and the ensemble of multiple networks can solve this problem.
According to the performance gap between \textbf{R} and \textbf{A}, it is demonstrated for the training scheme that trains each network on the different non-nodule samples to efficiently distinguish non-nodules from nodule candidates.
Furthermore, \textbf{AE} and \textbf{DAE} based on non-nodule categorization achieve state-of-the-art performances with low computational cost.
The main reason is that our non-nodule categorization effectively assists CNNs to learn representative features from a 2D single view.
In other words, CNNs can more easily learn meaningful features from the small number of similar non-nodules than from the large number of diverse non-nodules.

Although the proposed framework achieves state-of-the-art performance, there is still room for improvement.
First, 2D CNNs in our framework can be replaced with 3D CNNs appropriately constructed to learn representative features from 3D patches, if limited memory issue can be solved.
Second, our CNNs would be sensitive to the specific location of a candidate compared to multi-view CNNs or 3D CNNs due to the use of a 2D single view.
This limitation can be overcome by training on diversely located candidates and augmentation via translation.
In non-nodule categorization, we empirically selected the number of categories based on the balance between the number of nodule samples and the number of non-nodule samples.
By further studying how to adaptively select the optimal number of categories, we are able to improve our non-nodule categorization.
Also, non-nodules manually categorized by experts can make quantitative evaluation of non-nodule categorization results possible.

\section{Conclusion}
In this paper, we presented a novel false positive reduction framework, the ensemble of single-view 2D CNNs with fully automatic non-nodule categorization, in pulmonary nodule detection.
In contrast to recent frameworks employing 3D CNNs, our framework utilizes 2D CNNs based on 2D single views to improve computational efficiency.
Also, our training scheme using non-nodule categorization can extend the learning capability of representative features from diverse non-nodules.
We demonstrated the effectiveness of our framework through extensive experiments based on the dataset of the LUNA16 challenge.
Furthermore, we achieved state-of-the-art performance and low computational complexity compared to previous works.

\section*{Acknowledgment}
This work was supported by Institute for Information $\&$ communications Technology Promotion (IITP) grant funded by the Korea government (MSIT) (No.2017-0-01772 , Development of QA systems for Video Story Understanding to pass the Video Turing Test).

\section*{Conflict of interest}
The authors of this manuscript declare no conflict of interest.

\section*{References}

\bibliography{mybibfile}

\end{document}